\documentclass{article}

    \PassOptionsToPackage{numbers, compress}{natbib}

\usepackage[preprint]{neurips_data_2023}

\usepackage{tcolorbox}
\tcbuselibrary{skins,listings,breakable,poster}

\usepackage[utf8]{inputenc} 
\usepackage[T1]{fontenc}    
\usepackage{graphicx}
\usepackage{url}            
\usepackage{booktabs}       
\usepackage{amsfonts}       
\usepackage{nicefrac}       
\usepackage{microtype}      
\usepackage{xcolor}         
\usepackage{epsfig}
\usepackage{listings}
\usepackage{caption}
\usepackage{subcaption}
\usepackage{amsmath}
\usepackage{multirow}
\usepackage{array}
\usepackage{pifont}
\usepackage{hhline}
\usepackage{color, colortbl}
\usepackage{bm}
\usepackage{marvosym}
\usepackage{makecell}
\usepackage{adjustbox}
\usepackage{arydshln}
\usepackage{subfiles}
\usepackage{CJKutf8}
\usepackage{enumerate,letltxmacro}
\LetLtxMacro\itemold\item

\usepackage{amsfonts}       
\usepackage{nicefrac}       
\usepackage{microtype}      
\usepackage{epsfig}
\usepackage{listings}
\usepackage{caption}
\usepackage{subcaption}
\usepackage{amsmath}
\usepackage{multirow}
\usepackage{array}
\usepackage{pifont}
\usepackage{wrapfig}
\usepackage{hhline}
\usepackage{color, colortbl}
\usepackage{bm}
\usepackage{marvosym}
\usepackage{adjustbox}
\usepackage{arydshln}
\usepackage{makecell}
\usepackage{mmstyle}
\usepackage{threeparttable}
\usepackage{amssymb}
\usepackage{subfiles}
\usepackage{float}

\usepackage{rotating}
\usepackage{enumitem}
\usepackage{utfsym}
\usepackage[pagebackref,breaklinks,colorlinks,citecolor=eccvblue]{hyperref}
\usepackage[capitalize]{cleveref}

\newcommand{\tableCellHeight}{1}
\newcommand{\tabstyle}[1]{
  \setlength{\tabcolsep}{#1}
  \renewcommand{\arraystretch}{\tableCellHeight}
  \centering
  \small
}
\definecolor{improvement}{RGB}{225,97,78}
\definecolor{demphcolor1}{gray}{.6}
\definecolor{citecolor}{HTML}{0071bc}
\definecolor{tabhighlight}{HTML}{e5e5e5}

\crefname{section}{Sec.}{Secs.}
\Crefname{section}{Section}{Sections}
\Crefname{table}{Table}{Tables}
\crefname{table}{Tab.}{Tabs.}
\crefname{figure}{Fig.}{Figs.}
\Crefname{figure}{Fig.}{Figs.}
\crefname{appendix}{Appx.}{Appxs.}
\Crefname{appendix}{Appendix}{Appendixs}
\crefname{subsection}{Sec.}{Secs.}

\newcommand{\red}[1]{{\color{red}#1}}
\newcommand{\multilinecell}[1]{\renewcommand{\arraystretch}{1}\begin{tabular}{@{}c@{}}#1\end{tabular}}

\newcommand{\tablestyle}[2]{\setlength{\tabcolsep}{#1}\renewcommand{\arraystretch}{#2}\centering\footnotesize}
\newlength\savewidth\newcommand\shline{\noalign{\global\savewidth\arrayrulewidth
		\global\arrayrulewidth .8pt}\hline\noalign{\global\arrayrulewidth\savewidth}}

\definecolor{COLOR_MEAN}{HTML}{f0f0f0}
\definecolor{LIGHT_BLUE}{HTML}{cce4fe}
\definecolor{LIGHT_RED}{HTML}{f1b9b8}
\definecolor{LIGHT_YELLOW}{HTML}{f1f58a}
\newcommand{\MethodWord}{{21 }}

\newcommand{\graybox}[1]{\colorbox{gray!30}{#1}}

\newif\ifshowcomment
\showcommentfalse

\tcbset{
  aibox/.style={
    top=10pt,
    colback=white,
    colframe=black,
    colbacktitle=black,
    enhanced,
    center,
    attach boxed title to top left={yshift=-0.1in,xshift=0.15in},
    boxed title style={boxrule=0pt,colframe=white,},
  }
}
\newtcolorbox{AIbox}[2][]{aibox, title=#2,#1}
\usepackage[accsupp]{axessibility}

\usepackage{main_style}

\newcommand{\ourmethod}{{MMBench }}

\definecolor{COLOR_MEAN}{HTML}{f0f0f0}
\definecolor{LIGHT_BLUE}{HTML}{cce4fe}
\definecolor{LIGHT_RED}{HTML}{f1b9b8}
\definecolor{LIGHT_YELLOW}{HTML}{f1f58a}

\definecolor{COLOR_MEAN}{HTML}{f0f0f0}
\definecolor{LINK_COLOR}{HTML}{636EFA}
\hypersetup{
  colorlinks=true,
  linkcolor=LINK_COLOR,
  urlcolor=LINK_COLOR,
  citecolor=LINK_COLOR,
}

\title{MMBench: Is Your Multi-modal Model an All-around Player?}

\author{
  \hspace{-0.25cm}Yuan Liu$^{1, \ast}$, Haodong Duan$^{1,\ast,\ddagger}$, Yuanhan Zhang$^{2,\ast}$, Bo Li$^{2, \ast}$, \textbf{Songyang Zhang$^{1,\ast}$}, \\
  \hspace{-0.25cm}\textbf{ Wangbo Zhao$^{4}$, Yike Yuan$^{5}$, Jiaqi Wang$^{1}$, Conghui He$^{1}$, Ziwei Liu$^{2, \dagger}$, Kai Chen$^{1,\dagger}$} \\
  \hspace{-0.25cm}\textbf{Dahua Lin$^{1,3,\dagger}$} \\ 
 \hspace{-0.25cm}$^1$Shanghai AI Laboratory \quad $^2$Nanyang Technological University  \\
 \hspace{-0.25cm}$^3$ The Chinese University of Hong Kong \quad
 \hspace{-0.25cm}$^4$ National University of Singapore\\
 \hspace{-0.25cm}$^5$ Zhejiang University \\
 \hspace{-0.25cm}$^{\ast}$ Equal Contribution \quad $^{\ddagger}$ Project Lead \quad $^{\dagger}$ Corresponding Author \\
  }

\begin{document}
\maketitle
\begin{abstract}
\label{sec:abstract}
Large vision-language models (VLMs) have recently achieved remarkable progress, 
exhibiting impressive multimodal perception and reasoning abilities. 
However, effectively evaluating these large VLMs remains a major challenge,
hindering future development in this domain.
Traditional benchmarks like VQAv2 or COCO Caption provide quantitative performance measurements but lack fine-grained ability assessment and robust evaluation metrics.
Meanwhile, subjective benchmarks, such as OwlEval, offer comprehensive evaluations of a model's abilities by incorporating human labor, 
which is not scalable and may display significant bias.
In response to these challenges, we propose MMBench, a bilingual benchmark for assessing the multi-modal capabilities of VLMs.
MMBench methodically develops a comprehensive evaluation pipeline, primarily comprised of the following key features:
1. MMBench is meticulously curated with well-designed quality control schemes, surpassing existing similar benchmarks in terms of the number and variety of evaluation questions and abilities;
2. MMBench introduces a rigorous CircularEval strategy and incorporates large language models to convert free-form predictions into pre-defined choices,
which helps to yield accurate evaluation results for models with limited instruction-following capabilities.
3. MMBench incorporates multiple-choice questions in both English and Chinese versions, enabling an apples-to-apples comparison of VLMs' performance under a bilingual context.
To summarize, MMBench is a systematically designed \textbf{objective} benchmark for a \textbf{robust} and \textbf{holistic} evaluation of vision-language models. 
We hope MMBench will assist the research community in better evaluating their models and facilitate future progress in this area.  
The evalutation code of MMBench has been integrated into \href{https://github.com/open-compass/VLMEvalKit}{VLMEvalKit}~\cite{duan2024vlmevalkit}.
\footnote{This is a revised version released in April 2024. It describes MMBench v1.1, a refined version of the MMBench (with better data quality). Please refer to \url{https://arxiv.org/pdf/2307.06281v3} for the previous version, which is released in August 2023. }
\end{abstract}
\section{Introduction}
\label{sec:introduction}

\begin{figure}[!htbp]
\centering
\includegraphics[width=.8\linewidth]{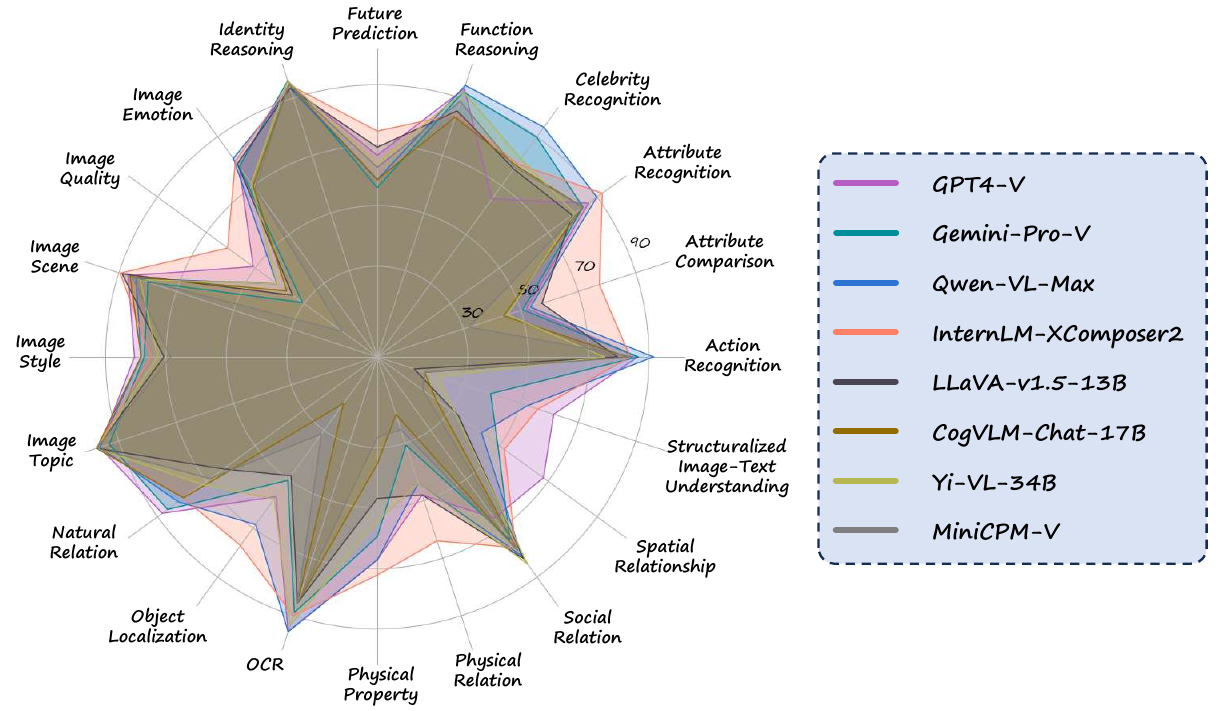}
\caption{\textbf{Results of eight representative large vision-language models (VLMs) across the 20 ability dimensions defined in MMBench-\texttt{test}.}}
\label{fig:0-ability-performance}
\vspace{-6mm}
\end{figure}

Recently, notable progress has been achieved within the realm of large language models (LLMs).
For instance, the latest LLMs, such as OpenAI's ChatGPT and GPT-4~\cite{OpenAI2023GPT4TR}, 
have demonstrated remarkable reasoning capabilities that are comparable to, 
and in some cases, even surpass human capabilities.
Drawing inspiration from these promising advancements in LLMs, 
large vision-language models (LVLMs) have also experienced a revolutionary transformation.
Notable works, such as GPT-4v~\cite{OpenAI2023GPT4TR}, Gemini-Pro-V~\cite{team2023gemini} and LLaVA~\cite{liu2023visual}, 
have demonstrated enhanced capabilities in image content recognition and reasoning within the domain of vision-language models,
exhibiting superior performance compared to earlier works.
Nevertheless, a large proportion of the early studies~\cite{gong2023multimodalgpt,zhu2023minigpt,liu2023visual} tend to emphasize showcasing qualitative examples rather than undertaking comprehensive and quantitative experiments to thoroughly assess their model performance.
The lack of quantitative assessment poses a considerable challenge for comparing various models. 
Recent studies have primarily explored two approaches to conduct quantitative evaluations.
The first approach involves utilizing existing public datasets~\cite{vqav2,coco_caption} for objective evaluation,
while the second approach employs human annotators~\cite{ye2023mplug,Xu2023LVLMeHubAC} to perform subjective evaluations.
However, it is worth noting that both approaches exhibit some inherent limitations.

A multitude of public datasets, such as VQAv2~\cite{vqav2}, COCO Caption~\cite{coco_caption}, GQA~\cite{hudson2019gqa}, and OK-VQA~\cite{ok-vqa}, 
have long served as valuable resources for the quantitative evaluation of VLMs. 
These datasets offer \textbf{objective} metrics, including accuracy, BLEU, CIDEr, \etc.
However, when employed to evaluate more advanced LVLMs, 
these benchmarks encounter the following challenges. 
1. \textbf{False Negative Issues}: 
Most existing evaluation metrics require an exact match between the prediction and the reference target, leading to potential limitations.
For instance, in the VQA task, even if the prediction is ``bicycle'' while the reference answer is ``bike'', 
the existing metric would assign a negative score to the prediction, 
resulting in a considerable number of false-negative samples.
2. \textbf{Lacking Finegrained Analysis}: 
Current public datasets predominantly focus on evaluating a model's performance on specific tasks, 
offering limited insights into the fine-grained capabilities of these models. 
Thus, they provide insufficient feedback regarding potential directions for future improvements.

Given the aforementioned challenges, recent studies, 
such as OwlEval~\cite{ye2023mplug} and LVLM-eHub~\cite{Xu2023LVLMeHubAC} propose human-involved \textbf{subjective} evaluation strategies, 
aiming to address existing methods' limitations by incorporating human judgment and perception in the evaluation process.
OwlEval artificially constructs 82 open-ended questions based on images from public datasets 
and employs human annotators to assess the quality of VLM predictions.
Similarly, inspired by FastChat~\cite{zheng2023judging}, 
LVLM-eHub develops an online platform where two models are prompted to answer the same question related to an image. 
A participant then compares the answers provided by two models. 
Subjective evaluation strategies offer numerous benefits.
These include \textbf{accurate matching}, where humans can precisely correlate a prediction with the target, even when expressed in different words, 
and \textbf{comprehensive assessment}, where humans are inclined to juxtapose two predictions considering multiple facets. 
The ultimate score is computed as the mean score across diverse abilities, facilitating a holistic evaluation of the model's capabilities.

While subjective evaluation allows for a more comprehensive assessment of a VLM,
it also introduces new challenges.
Firstly, human evaluations are inherently biased.
Consequently, it becomes challenging to reproduce the results presented in a work with a different group of annotators.
Also, existing subjective evaluation strategies face scalability issues.
Employing annotators for model evaluation after each experiment is an expensive endeavor.
Moreover, evaluation datasets of small sizes can result in statistical instability.
To ensure a robust evaluation, collecting more data becomes necessary, which in turn demands a significant amount of human labor.

In light of the challenges faced by conventional objective and subjective benchmarks, 
we propose \textbf{MMBench}, a systematically designed objective evaluation benchmark to robustly evaluate different abilities of large vision-language models. 
Currently, MMBench contains over 3000 multiple-choice questions covering 20 different ability dimensions, 
such as object localization and social reasoning, for evaluating vision-language models. 
Each ability dimension encompasses over 125 questions, 
with the quantity of questions per ability maintained at a roughly equal level. 
The distribution facilitates a balanced and thorough assessment of these abilities. 
Since some existing VLMs have limited instruction-following capability and cannot directly output choice labels (A, B, C, \emph{etc.}) for MMBench questions, 
the evaluation based on exact matching may not yield accurate and reasonable conclusions. 
In order to reduce the number of false-negative samples during answer matching, 
we employ GPT-4 to match a model's prediction to candidates choices in a multi-choice question and then output the label for the matched choice. 
We conduct a comparison between GPT-4-based choice matching and human evaluations, and discovered that GPT-4 can accurately match human assessments in 91.5\% of cases,
demonstrating its good alignment and robustness as a choice extractor. 
To make the evaluation more robust, we propose a novel evaluation strategy, named \textbf{CircularEval} (details in \Cref{subsec:eval_strategy}). 
We comprehensively evaluate \MethodWord well-known vision-language models (across different model architectures and scales) on MMBench and report their performance on different ability dimensions. 
The performance ranking offers a direct comparison between various models and provides valuable feedback for future optimization. 
In summary, our main contributions are three-fold:

\begin{enumerate}[label={\bf {{$\bullet$}}},,leftmargin=*,topsep=0.5ex,itemsep=-0.5ex,partopsep=0.75ex,parsep=0.75ex,partopsep=0pt,wide,labelindent=0pt]
    \item \textbf{Systematically-constructed Dataset}: To thoroughly evaluate the capacity of a VLM,
    we carefully curated a dataset comprising a total of 3,217 meticulously selected questions, covering a diverse spectrum of 20 fine-grained skills.
    \item \textbf{Robust Evaluation}: We introduce a novel circular evaluation strategy (CircularEval) to improve the robustness of our evaluation process. 
    After that, GPT-4 is employed to match the model's prediction with given choices, which can successfully extract choices even from predictions of a VLM with poor instruction-following capability.
    \item \textbf{Analysis and Observations}: We perform a comprehensive evaluation of a series of well-known vision-language models using MMBench, and the evaluation results can provide insights to the research community for future improvement.
\end{enumerate}

\section{Related Work}
\label{sec:related_work}

\subsection{Multimodal Datasets}
Large-scale VLMs have shown promising potential in multimodal tasks such as complex scene understanding and visual question answering. 
Though qualitative results so far are encouraging, 
quantitative evaluation is of great necessity to systematically evaluate and compare the abilities of different VLMs.
Recent works have evaluated their models on numerous existing public multi-modality datasets.
COCO Caption~\cite{coco_caption}, Nocaps~\cite{nocaps}, and Flickr30k~\cite{flickr30k} provide human-generated image captions and the corresponding task is to describe the image content in the form of text.
Visual question answering datasets, such as GQA~\cite{hudson2019gqa}, OK-VQA~\cite{ok-vqa}, VQAv2~\cite{vqav2}, and Vizwiz~\cite{vizwiz}, 
contain question-answer pairs related to the given image, 
used to measure the model's ability on visual perception and reasoning.
Some datasets provide more challenging question-answering scenarios by incorporating additional tasks. 
For example, TextVQA~\cite{textvqa} proposes questions about text shown in the image, thus involving the OCR task in question-answering.
ScienceQA~\cite{scienceqa} focuses on scientific topics, requiring the model to integrate commonsense into reasoning.
Youcook2~\cite{youcook2} replaces images with video clips, introducing additional temporal information.
However, the aforementioned datasets are designed on specific domains, and can only evaluate the model's performance on one or several tasks. 
Besides, different data formats and evaluation metrics across datasets make it more difficult to comprehensively assess a model's capability.
Ye et al.~\cite{ye2023mplug} constructed OwlEval, an evaluation set encompassing a variety of visual-related tasks, albeit of a limited size. 
Fu et al.~\cite{Fu2023MMEAC} introduced MME, which assesses a VLM's capabilities from various perspectives at a small scale. 
Diverging from prior works, in this paper, we present a novel multimodal benchmark, MMBench. 
We also devise a suite of evaluation standards aimed at ensuring the stability and accuracy of the evaluation results.

\subsection{Multimodal Models}
Building upon the success of Large Language Models (LLMs) such as GPTs~\cite{radford2019language, brown2020language, ouyang2022training}, LLaMA~\cite{touvron2023llama}, and Vicuna~\cite{zheng2023judging}, 
recent advancements have been made in multimodal models. Flamingo~\cite{alayrac2022flamingo}, an early attempt at integrating LLMs into vision-language pretraining, has made significant strides. 
To condition effectively on visual features, it incorporates several gated cross-attention dense blocks within pretrained language encoder layers. 
OpenFlamingo~\cite{alayrac2022flamingo} offers an open-source version of this model. 
BLIP-2~\cite{li2023blip} introduces a Querying Transformer (Q-former) to bridge the modality gap between the frozen image encoder and the large language encoder. 
Subsequently, InstructBLIP~\cite{dai2023instructblip} extends BLIP-2~\cite{li2023blip} with vision-language instruction tuning, achieving superior performance. 
MiniGPT-4~\cite{zhu2023minigpt} attributes the prowess of GPT-4~\cite{OpenAI2023GPT4TR} to advanced LLMs and proposes the use of a single projection layer to align the visual representation with the language model. 
LLaVA~\cite{liu2023visual} also utilizes GPT-4 to generate instruction-following data for vision-language tuning. 
The learning paradigm and the multimodal instruction tuning corpus proposed by LLaVA are widely adopted by subsequent works~\cite{liu2023improved,chen2023sharegpt4v,2023YiVL,2023xtuner}.
During the instruction tuning, Low-Rank Adaptation (LoRA~\cite{hu2021lora}) has been adopted by recent works~\cite{ye2023mplug,internlmxcomposer2,2023xtuner} on language models to achieve better performance on multimodal understanding.
In the realm of proprietary models, the APIs of multiple powerful VLMs have also been made publicly available to prosper downstream applications, including GPT-4v~\cite{OpenAI2023GPT4TR}, Gemini-Pro-V~\cite{team2023gemini}, and Qwen-VL-Max~\cite{bai2023qwen}. 
After conducting a thorough evaluation of these models on the proposed MMBench, we offer insights for future multimodal research.
\section{The construction of MMBench}

\label{sec:mmagibenchmark}
Three characteristics differentiate MMBench from existing benchmarks for multi-modality understanding:
i) MMBench adopts images / problems from various sources to evaluate diversified abilities in a hierarchical taxonomy;
ii) MMBench performs rigorous quality control to ensure the correctness and validity of testing samples; 
iii) MMBench is a bilingual multi-modal benchmark and enables an apple-to-apple comparison of VLM performance under English and Chinese contexts.
Below we will delve into more details of the construction of MMBench.

\begin{figure}[t]
\centering
\includegraphics[width=0.8\linewidth]{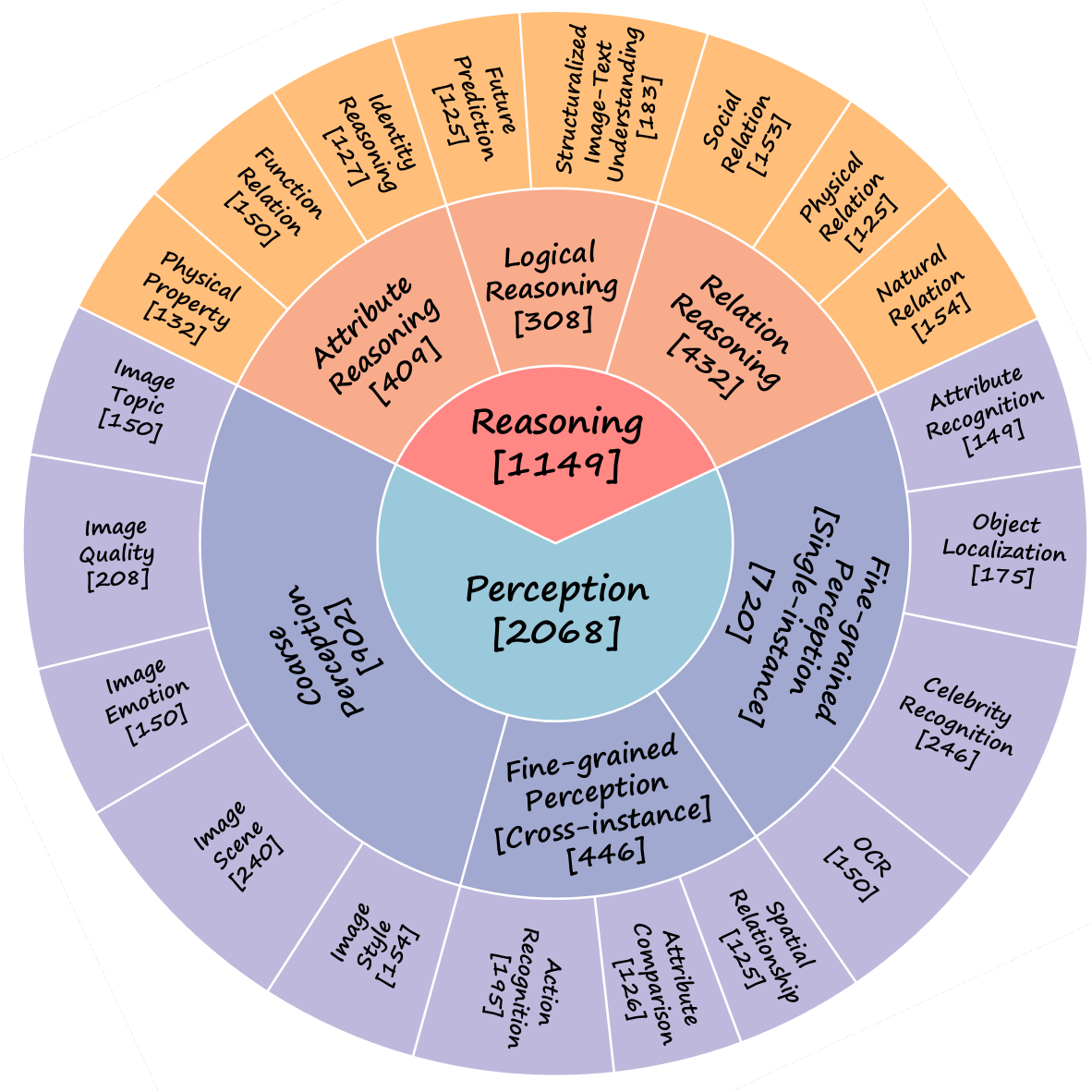}
\caption{\textbf{Ability dimensions in MMBench.} 
Currently, \ourmethod incorporates three levels of ability dimensions, encompassing 20 distinct leaf abilities. }
\label{fig:1-mm_ability}
\end{figure}

\subsection{The Hierachical Ability Taxonomy of \ourmethod}
\label{method:ability_dim}
Human possess remarkable perception and reasoning capabilities. 
These abilities have been crucial in human evolution and serve as a foundation for complex cognitive processes. 
Perception refers to gathering information from sensory inputs, 
while reasoning involves drawing conclusions based on this information. 
Together, they form the basis of most tasks in the real world, including recognizing objects, solving problems, and making decisions~\cite{oaksford2007bayesian,fodor1983modularity}. 
In pursuit of genuine general artificial intelligence (AGI), vision-language models (VLMs) are also expected to exhibit strong perception and reasoning abilities. 
Therefore, we adopt \textbf{Perception} and \textbf{Reasoning} as level-1 (\textbf{L-1}) abilities in our taxonomy. 
After that, we incorporate more fine-grained ability dimensions into the taxonomy, and categorize them into six \textbf{L-2} and twenty \textbf{L-3} ability dimensions. 
We display the ability taxonomy in \Cref{fig:1-mm_ability} and 
you can find detailed definitions of each fine-grained ability in the Appendix.

\subsection{Data Collection and Quality Control}
\label{method:collection}

\noindent \textbf{Question Collection. }
In MMBench, we collect vision-language QAs in the format of multiple-choice problems for each L-3 ability.
A problem $P_i$ corresponds to a quadruple $(Q_i, C_i, I_i, A_i)$. 
$Q_i$ denotes the question, 
$C_i$ represents a set with $n$ ($2\leq n\leq 4$) choices $c_1, c_2, ..., c_n$, 
$I_i$ corresponds to the image associated with the question, 
and $A_i$ is the correct answer.
The data --- including images, choices, and questions --- 
are manually collected from multiple sources by a group of volunteers. 
For each \textbf{L-3} ability, we first set an example by compiling $10\sim 20$ multiple-choice questions. 
Then we enlist the volunteers, all of whom are undergraduate or graduate students from various disciplines, to expand the problem set.
The expansion is based on the ability definition and potential data sources, which include both public datasets and the Internet. According to the statistics,
more than 80\% of questions in MMBench are collected from the Internet.
For the remaining 20\% samples, the images are gathered from the validation set of public datasets (if they exist) while the questions are self-constructed, which is not supposed to be used for training. 
In the Appendix, we list data sources used in collection and provide visualization of samples corresponding to each \textbf{L-3} ability.

\begin{figure}[t]
    \centering
    \includegraphics[width=\textwidth]{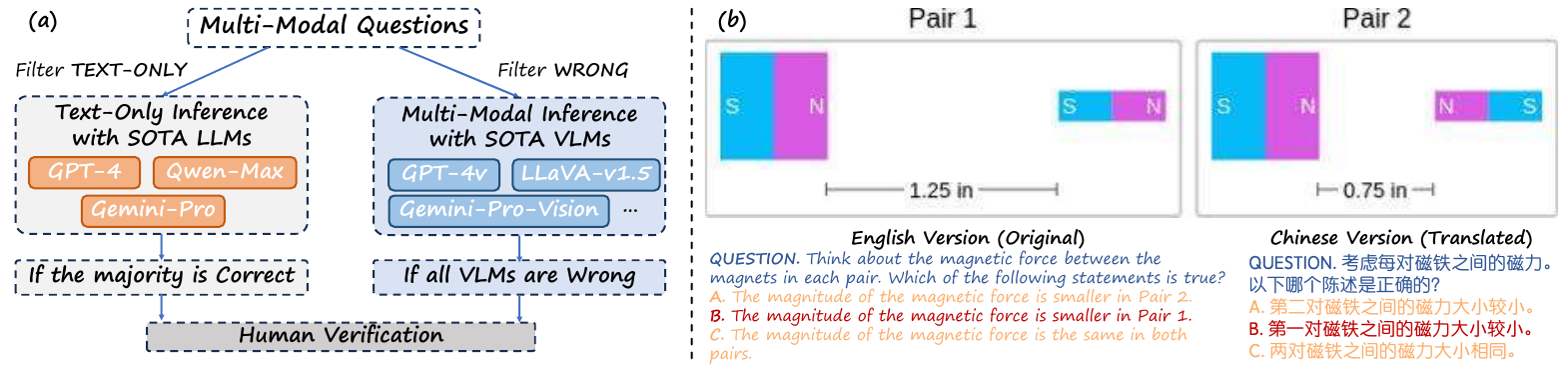}
    \caption{\textbf{The construction of MMBench. } (a). The quality control  strategies adopted in MMBench; (b) An illustration of questions in MMBench-CN.}
    \label{fig:bench_construct}
\end{figure}

\noindent \textbf{Quality Control. }
Raw data collected from volunteers may include wrong or unqualified samples. 
During investigation, we find that there exist two major patterns for such samples:
i) the answer to the question can be inferred with \textbf{text-only} inputs, which makes it inappropriate for evaluating the multimodal understanding capability of VLMs;
ii) the sample is simply \textbf{wrong}, either with a flawed question, choices, or an incorrect answer.
We design two strategies to filter those low-quality samples, which is visualized in \Cref{fig:bench_construct}(a). 
We adopt `majority voting' to detect \textbf{text-only} samples:
data samples are inferred with state-of-the-art LLMs (GPT-4~\cite{OpenAI2023GPT4TR}, Gemini-Pro~\cite{team2023gemini}, \textit{etc.}). 
If more than half of the LLMs can answer the question correctly with text-only inputs, 
the question will be manually verified and then removed if it is unqualified. 
To detect \textbf{wrong} samples, we also implement an automatic filtering mechanism. 
We select several state-of-the-art VLMs (including both open-source and proprietary ones), 
to answer all questions in \ourmethod. 
If all VLMs fail to answer the question correctly, we consider this question potentially problematic. 
Such questions will be manually checked and excluded if they are actually wrong.
The quality control paradigm helps us to construct high-quality datasets and can also be used to clean other existing benchmarks.

\noindent \textbf{MMBench-CN. } 
We further convert the curated MMBench into a Chinese version. 
During the process, all content in questions and choices are translated to Chinese based on GPT-4, except for proper nouns, symbols, and code. 
All those translations are verified by humans to ensure the validity. 
MMBench-CN enables an apple-to-apple comparison of VLM performance under English and Chinese contexts.
An example in MMBench-CN is illustrated in \Cref{fig:bench_construct}(b).

\subsection{MMBench Statistics }

\noindent \textbf{Data Statistics.} In the present study, we have gathered a total of 3,217 data samples spanning across 20 distinct \textbf{L-3} abilities. 
We depict the problem counts of all the 3 levels of abilities in \Cref{fig:1-mm_ability}. 
To ensure a balanced and comprehensive evaluation for each ability, 
we try to maintain an even distribution among problems associated with different abilities during data collection, 
with at least 125 samples for each \textbf{L-3} category. 

\noindent \textbf{Data Splits.} We follow the standard practice in previous works~\cite{ok-vqa} to split \ourmethod into \texttt{dev} and \texttt{test} subsets at a ratio of 4:6. 
For the \texttt{dev} subset, we make all data samples publicly available along with the ground truth answers for all questions. 
For the \texttt{test} subset, only the data samples are released, while the ground truth answers remain confidential.
To obtain the test subset evaluation results, one needs to submit the predictions to \ourmethod evaluation server.

\section{Evaluation Strategy}
\label{method:eval_strategy}

\begin{figure}[t]
    \centering
    \includegraphics[width=\textwidth]{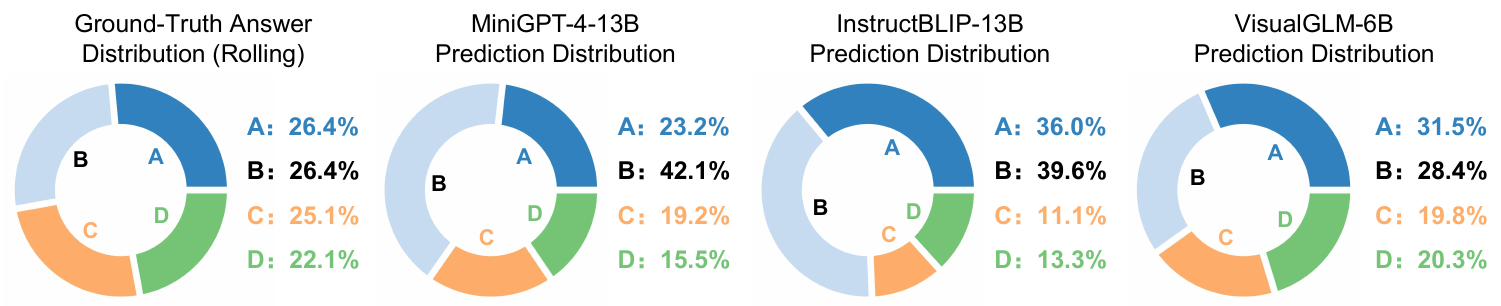}
    \caption{\textbf{The choice distribution of ground-truth answers and predictions of sample VLMs (all \textit{CircularEval} records). } 
    Since there exist questions with only 2/3 choices in MMBench, the choice distribution of ground-truth is not exactly even. }
    \label{fig:ABCD}
\end{figure}

In MMBench, we propose a new strategy that yields robust evaluation results with affordable costs. 
To deal with the free-form outputs of VLMs, 
we propose utilizing state-of-the-art LLMs as a helper for choice extraction.
We conduct extensive experiments to study the LLM-involved evaluation procedure. 
The results well support the effectiveness of GPT-4 as a choice extractor. 
We further adopt a new evaluation strategy named \textbf{CircularEval}, 
which feeds a question to a VLM multiple times (with shuffled choices) 
and checks if a VLM succeeds in all attempts.
With \textbf{CircularEval}, we deliver a rigorous evaluation and more effectively display the performance gap between VLMs.

\subsection{LLM-involved Choice Extraction }
\label{subsec:choice_extraction_main}

In our initial attempts to evaluate on MMBench questions,
we observed that the instruction-following capabilities of VLMs can vary significantly.
Though problems are presented as clear multiple-choice questions with well-formatted options, many VLMs still output the answers in free-form text\footnote{For example, 
the model output can be the \graybox{meaning of choice ``A''} rather than \graybox{``A''}. }, 
especially for VLMs that have not been trained with multiple-choice questions or proprietary VLMs for general purposes (GPT-4v, Qwen-VL-Max, \emph{etc.}). 
Extracting choices from free-form predictions is straight-forward for human beings, but might be difficult with rule-based matching. 
To this end, we design a universal evaluation strategy for all VLMs with different instruction-following capabilities: 

\noindent \textbf{Step 1. Matching Prediction. } 
Initially, we attempt to extract choices from VLM predictions using heuristic matching. 
We aim to extract the choice label (e.g., A, B, C, D) from the VLM's output. 
If successful, we use this as the prediction. If not, we attempt to extract the choice label using an LLM.

\noindent \textbf{Step 2. Matching LLM's output. }  
If step 1 fails, we try to extract the choice with LLMs (\textbf{gpt-4-0125} by default). 
We first provide ChatGPT with the question, choices, and model prediction.
Then, we request it to align the prediction with one of the given choices, and subsequently produce the label of the corresponding option. 
If the LLM finds that the model prediction is significantly different from all choices, we ask it to return a pseudo choice `Z'. 
In experiments, we find that for almost all cases we encountered, 
the LLM can output a valid choice according to the instruction. 
For each sample, we compare the model's label prediction (after GPT's similarity readout) with the actual ground truth label. 
If the prediction matches the label, the test sample is considered correct.

\subsection{LLM as the Choice Extractor: A Feasibility Analysis}
\label{sec:feasibility}

\noindent \textbf{Instruction following (IF) capabilities of VLMs vary a lot. }
We conduct pilot experiments to study the effectiveness of LLMs as the choice extractor.
As a first step, we perform single-pass inference on all MMBench questions with VLMs in our evaluation core set (defined in \Cref{sec:main_results}). 
While there exist VLMs that perfectly follow the multiple-choice format and achieve high success rates ($>99\%$) in heuristic matching, 
all proprietary models and a significant proportion of open-source VLMs failed to generate well-formatted outputs. 
In \Cref{tab:matching_rate}, we list the success rates of different VLMs in heuristic matching\footnote{
VLMs that achieve $>99\%$ matching rates are not listed, including LLaVA series, Yi-VL series, mPLUG-Owl2, OpenFlamingo v2, and CogVLM-Chat. }. 
Among all VLMs, VisualGLM achieves the lowest matching success rate, which is merely 65\%. 
For those VLMs, incorporating LLMs as the choice extractor leads to significant change in the final accuracy. 
Another noteworthy thing is that
the IF capability and the overall multimodal understanding capability is not necessarily correlated. 
For example,
OpenFlamingo v2~\cite{alayrac2022flamingo} demonstrates top IF capability among all VLMs, while also achieving one of the worst performances on MMBench (\Cref{tab:main_results_short}). 

\noindent
\begin{table}[t]
\begin{minipage}{0.62\linewidth}
\captionof{table}{\textbf{Statistics of IF capabilities of VLMs. } 
We report the heuristic matching success rate of VLMs, and the accuracy before and after LLM-based choice extraction.
In `X\red{+Y}', X denotes the matching-based accuracy, \red{Y} indicates the gain of using LLM as the choice extractor. }
\label{tab:matching_rate}
\centering
\resizebox{\linewidth}{!}{
\tablestyle{3pt}{1.6}
\begin{tabular}{lcclcc}
\shline
Model Name          & Match Rate & DEV Acc  & Model Name & Match Rate & DEV Acc  \\ \shline
MiniGPT4-7B         & 85.7 & 47.9 \red{+8.8}  & MiniGPT4-13B    & 84.8 & 52.1 \red{+8.7}  \\
InstructBLIP-7B     & 93.6 & 57.1 \red{+4.3}  & InstuctBLIP-13B & 93.7 & 58.4 \red{+5.6}  \\
IDEFICS-9B-Instruct & 96.6 & 58.4 \red{+1.5}  & Qwen-VL-Chat    & 93.8 & 73.3 \red{+3.6}  \\
MiniCPM-V           & 95.2 & 70.9 \red{+4.5}  & VisualGLM-6B    & 64.8 & 39.9 \red{+23.2} \\ \shline
GPT-4v              & 91.8 & 81.5 \red{+3.6}  & GeminiProVision & 97.5 & 81.8 \red{+0.8}  \\
Qwen-VL-Plus        & 77.4 & 64.5 \red{+15.0} & Qwen-VL-Max     & 96.0 & 82.0 \red{+3.2} \\ \shline
\end{tabular}
}
\end{minipage}
\hfill
\begin{minipage}{0.35\linewidth}
\centering
\includegraphics[width=.85\linewidth]{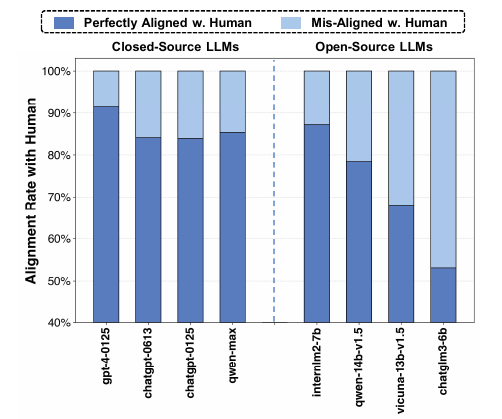}
\captionof{figure}{\textbf{Alignment rates between human and different LLMs. } `chatgpt' is `gpt-3.5-turbo'. Open-source LLMs are `chat' variants. }
\label{fig:alignment}
\end{minipage}
\vspace{-5mm}
\end{table}

\begin{figure}[t]
    \vspace{3mm}
    \centering
    \includegraphics[width=\textwidth]{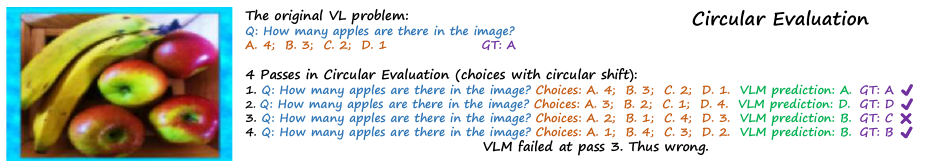}
    \caption{\textbf{CircularEval strategy.} In \textbf{CircularEval}, a problem is tested multiple times with circular shifted choices and 
    the VLM needs to succeed in all testing passes. In this example, the VLM failed in pass 3 and thus considered failed the problem. }
    \label{fig:circular}
\end{figure}

\noindent \textbf{Quality and stability of LLM Choice Extractors. }
For VLM predictions that cannot be parsed by heuristic matching, we adopt GPT-4 as the choice extractor. 
To validate its efficacy, we first build a subset of the inference records. 
Each item in the set is a pair of questions and VLM predictions, which cannot be parsed by step-1 matching. 
We sample 10\% of those hard examples ($\sim$ 420 samples), 
and ask volunteers to perform manual choice extraction on these data samples.
Such annotations enable us to validate the choice extraction of LLMs, by measuring their alignment rates with humans.

\Cref{fig:alignment} reports the alignment rates (extracted choices are exactly the same) between LLMs and humans. 
We find that a great number of LLMs can complete the task well and achieve decent alignment rate with human. 
Among proprietary LLMs, GPT-4 achieves the highest level of alignment rate, which is 91.5\%, 
while GPT-3.5-Turbo and Qwen-Max achieve around 85\%. 
Open-source LLMs achieve more diversified performance on the choice matching task.
InternLM2-7B~\cite{2023internlm} achieves an 87\% alignment rate and significantly outperforms other open-source LLMs and GPT-3.5-Turbo.
In the following experiments, we adopt \textbf{gpt-4-0125} as the choice extractor due to its superior alignment capability.
Meanwhile, we also note that the slight difference in top-performing LLMs' alignment rates has little effect on the quantitative performance of VLMs.

\subsection{CircularEval Strategy }
\label{subsec:eval_strategy}
In MMBench, the problems are presented as multiple-choice questions. 
Such formulation poses an evaluation challenge: 
random guessing will lead to $\sim$25\% Top-1 accuracy for 4-choice questions, 
potentially reducing the discernible performance differences among VLMs. 
Besides, we noticed that VLMs may prefer to predict a certain choice among all given choices (\Cref{fig:ABCD}), which further amplifies the bias in evaluation. 
To this end, we introduce a more robust evaluation strategy termed \textbf{Circular Evaluation} (or \textbf{CircularEval}). 
Under this setting, each question is fed to a VLM $N$ times ($N$ is the number of choices).
Each time, circular shifting is applied to the choices and the answer to generate a new prompt for VLMs (example in \Cref{fig:circular}).
A VLM is considered successful in solving a question only if it correctly predicts the answer in all circular passes. 
In practice, once a VLM fails on a circular passes,
there is no need to infer the remaining passes, 
which makes the actual cost of CircularEval less than $N\times$ under practical scenarios. 
CircularEval can achieve a good trade-off between robustness and cost.
\section{Evaluation Results}
\label{sec:exp}

\subsection{Experimental Setup}
For the main results, 
we evaluate various models belonging to three major categories on MMBench: 
(a) \textit{Text-Only} GPT-4~\cite{OpenAI2023GPT4TR}; 
(b) \textit{Open-Source VLMs} including model variants of OpenFlamingo~\cite{alayrac2022flamingo}, MiniGPT4~\cite{zhu2023minigpt}, InstructBLIP~\cite{dai2023instructblip}, LLaVA~\cite{liu2023improved}, IDEFICS~\cite{laurencon2023obelics}, CogVLM~\cite{Wang2023CogVLMVE}, Qwen-VL~\cite{bai2023qwen}, Yi-VL~\cite{2023YiVL}, mPLUG-Owl~\cite{Ye2023mPLUGOwl2RM}, InternLM-XComposer~\cite{internlmxcomposer2}, and MiniCPM-V~\cite{minicpm2024};
(c) \textit{Proprietary VLMs} including Qwen-VL-[Plus/Max]~\cite{bai2023qwen}, Gemini-Pro-V~\cite{team2023gemini}, and GPT-4v~\cite{OpenAI2023GPT4TR}.
For a fair comparison, we adopt the zero-shot setting to infer MMBench questions with all VLMs, based on the same prompt. 
For all VLMs, open-ended generation is adopted to obtain the prediction,
and `\textbf{gpt-4-0125}' is used as the choice extractor. 
In the Appendix, we provide detailed information regarding the architecture and the parameter size for all Open-Source VLMs evaluated in this paper, 
as well as additional results for more VLMs under various settings. 
We conduct all the evaluation with VLMEvalKit~\cite{duan2024vlmevalkit}. 

\subsection{Main Results}
\label{sec:main_results}

\begin{table}[t]
\centering
\caption{\textbf{CircularEval $vs.$ VanillaEval. } 
We report the \textbf{CircularEval} Top-1 accuracy and accuracy drop (compared to \textbf{VanillaEval}) of all VLMs on MMBench-\texttt{dev}. 
}  
\vspace{2mm}
\label{tab:circular_vs_vanilla}
\resizebox{\textwidth}{!}{%
\tablestyle{4pt}{1.5}
\begin{tabular}{lcc|lcc|lcc}
\shline
\rowcolor{COLOR_MEAN}
\textbf{VLM} & \textbf{Circular} & \textbf{Acc Change} & \textbf{VLM} & \textbf{Circular} & \textbf{Acc Change} & \textbf{VLM} & \textbf{Circular} & \textbf{Acc Change} \\ 
\textbf{MiniGPT4-7B} & 32.7\% & \red{\textbf{-24.1\%}} & \textbf{MiniGPT4-13B} & 37.5\% & \red{\textbf{-23.2\%}} & \textbf{Yi-VL-6B} & 65.6\% & \red{\textbf{-9.8\%}} \\ 
\textbf{InstructBLIP-7B} & 37.4\% & \red{\textbf{-24.0\%}} & \textbf{InstructBLIP-13B} & 40.9\% & \red{\textbf{-23.0\%}} & \textbf{Yi-VL-34B} & 68.2\% & \red{\textbf{-9.5\%}} \\ 
\textbf{LLaVA-v1.5-7B} & 62.5\% & \red{\textbf{-11.2\%}} & \textbf{LLaVA-v1.5-13B} & 67.2\% & \red{\textbf{-8.6\%}} & \textbf{MiniCPM-V} & 64.8\% & \red{\textbf{-10.6\%}} \\ 
\textbf{IDEFICS-9B-Instruct} & 37.2\% & \red{\textbf{-22.6\%}} & \textbf{LLaVA-InternLM2-20B} & 72.8\% & \red{\textbf{-7.0\%}} & \textbf{Qwen-VL-Plus} & 62.9\% & \red{\textbf{-16.6\%}} \\ 
\textbf{VisualGLM-6B} & 36.1\% & \red{\textbf{-27.0\%}} & \textbf{CogVLM-Chat-17B} & 62.4\% & \red{\textbf{-15.6\%}} & \textbf{Qwen-VL-Max} & 76.4\% & \red{\textbf{-8.7\%}} \\ 
\textbf{Qwen-VL-Chat} & 59.5\% & \red{\textbf{-17.4\%}} & \textbf{mPLUG-Owl2} & 63.5\% & \red{\textbf{-8.7\%}} & \textbf{Gemini-Pro-V} & 70.9\% & \red{\textbf{-11.7\%}} \\ 
\textbf{OpenFlamingo v2} & 2.6\% & \red{\textbf{-34.1\%}} & \textbf{InternLM-XComposer2} & 79.1\% & \red{\textbf{-4.7\%}} & \textbf{GPT-4v} & 74.3\% & \red{\textbf{-10.8\%}} \\ 
\shline
\end{tabular}}
\vspace{-4mm}
\end{table}

\noindent \textbf{CircularEval \textit{vs.} VanillaEval. }
Before delving deeper into concrete evaluation results, we first compare our \textbf{CircularEval} (infer a question over multiple passes, consistency as a must) with \textbf{VanillaEval} (infer a question only once). 
In \Cref{tab:circular_vs_vanilla}, we present the results with two evaluation strategies on MMBench-\texttt{dev}. 
For most VLMs, switching from VanillaEval to CircularEval leads to a significant drop in model accuracy.
In general, comparisons under CircularEval can reveal a more significant performance gap between different VLMs.
LLaVA-v1.5-13B outperforms its 7B counterpart by 2.1\% Top-1 accuracy under VanillaEval, while a much larger performance gap (4.7\% Top-1) is observed under CircularEval.
As a special case, the performance of OpenFlamingo v2 drops from 36.7\% to only 2.6\% when we move from VanillaEval to CircularEval.
CircularEval is such a challenging setting that it even makes state-of-the-art proprietary VLMs (GPT-4v, Qwen-VL-Max, \emph{etc.}) suffer from $\sim$10\% Top-1 accuracy drops.
In the following experiments, we adopt the more rigorous and well-defined \textbf{CircularEval} as our default evaluation paradigm.

We exhaustively evaluate all VLMs on all existing leaf abilities of MMBench. 
In \Cref{tab:main_results_short}, we report the models' overall performance and the performance in six \textbf{L-2} abilities on the \texttt{test} split, namely 
Coarse Perception (\textbf{CP}), Fine-grained Perception (single-instance, \textbf{FP-S}; cross-instance, \textbf{FP-C}), Attribute Reasoning (\textbf{AR}), Logic Reasoning (\textbf{LR}), and Relation Reasoning (\textbf{RR}).\footnote{
Please refer to the appendix for more fine-grained results and MMBench-\texttt{dev} split results. }
The results offer valuable insights into the individual strengths and limitations of each VLM in multi-modal understanding.

\begin{table}[t]
\centering
\caption{\textbf{CircularEval results on MMBench \texttt{test} set (L-2 abilities).} 
Abbreviations adopted: LR for Logical Reasoning; AR for Attribute Reasoning; RR for Relation Reasoning; FP-C for Fine-grained Perception (Cross Instance); FP-S for Fine-grained Perception (Single Instance); CP for Coarse Perception. Models are sorted by the ascending order of overall accuracy (intra-group). 
Open-source models tagged with * incorporate in-house data in model training. } 
\vspace{2mm}
\label{tab:main_results_short}
\resizebox{\textwidth}{!}{%
\tablestyle{10pt}{1.5}
\begin{tabular}{lccccccc}
\shline
\textbf{Model} & \textbf{Overall} & \textbf{CP} & \textbf{FP-S} & \textbf{FP-C} & \textbf{AR} & \textbf{LR} & \textbf{RR} \\ \shline 
\rowcolor{LIGHT_YELLOW}
\multicolumn{8}{c}{\textbf{Large Language Models}} \\
\textbf{GPT-4-Turbo (0125)}~\cite{OpenAI2023GPT4TR} & 2.9\% & 0.6\% & 1.2\% & 4.1\% & 3.7\% & 4.9\% & 7.4\% \\ 
\shline
\rowcolor{LIGHT_BLUE}
\multicolumn{8}{c}{\textbf{OpenSource VLMs}} \\
\textbf{OpenFlamingo v2}~\cite{alayrac2022flamingo} & 2.3\% & 1.1\% & 3.5\% & 1.5\% & 5.3\% & 0.0\% & 2.7\% \\ 
\textbf{MiniGPT4-7B}~\cite{zhu2023minigpt} & 30.5\% & 37.0\% & 31.8\% & 17.2\% & 49.8\% & 9.2\% & 25.6\% \\ 
\textbf{IDEFICS-9B-Instruct}~\cite{laurencon2023obelics} & 35.2\% & 48.3\% & 31.3\% & 29.6\% & 47.8\% & 11.4\% & 25.2\% \\ 
\textbf{VisualGLM-6B}~\cite{du2022glm} & 35.4\% & 40.2\% & 38.5\% & 26.2\% & 47.8\% & 19.6\% & 29.5\% \\ 
\textbf{InstructBLIP-7B}~\cite{dai2023instructblip} & 38.3\% & 46.7\% & 39.0\% & 31.8\% & 55.5\% & 8.7\% & 31.0\% \\ 
\textbf{MiniGPT4-13B}~\cite{zhu2023minigpt} & 38.8\% & 44.6\% & 42.9\% & 23.2\% & 64.9\% & 8.2\% & 32.9\% \\ 
\textbf{InstructBLIP-13B}~\cite{dai2023instructblip} & 39.8\% & 47.2\% & 42.9\% & 21.0\% & 60.4\% & 12.5\% & 38.8\% \\ 
\textbf{Qwen-VL-Chat*}~\cite{bai2023qwen} & 60.9\% & 68.5\% & 67.7\% & 50.2\% & 78.0\% & 37.0\% & 45.7\% \\ 
\textbf{MiniCPM-V}~\cite{minicpm2024} & 61.4\% & 65.6\% & 69.4\% & 51.3\% & 70.6\% & 35.3\% & 59.7\% \\ 
\textbf{LLaVA-v1.5-7B}~\cite{liu2023improved} & 63.4\% & 70.0\% & 68.0\% & 57.7\% & 77.6\% & 33.2\% & 56.2\% \\ 
\textbf{mPLUG-Owl2}~\cite{Ye2023mPLUGOwl2RM} & 63.5\% & 68.1\% & 69.1\% & 55.8\% & 78.4\% & 37.0\% & 57.0\% \\ 
\textbf{CogVLM-Chat-17B}~\cite{Wang2023CogVLMVE} & 63.6\% & 72.8\% & 66.6\% & 55.4\% & 71.4\% & 33.7\% & 62.0\% \\ 
\textbf{Yi-VL-6B*}~\cite{2023YiVL} & 65.5\% & 72.8\% & 72.9\% & 56.2\% & 75.5\% & 41.3\% & 55.4\% \\ 
\textbf{LLaVA-v1.5-13B}~\cite{liu2023improved} & 66.9\% & 73.1\% & 72.4\% & 60.3\% & 75.5\% & 35.9\% & 65.5\% \\ 
\textbf{Yi-VL-34B*}~\cite{2023YiVL} & 68.4\% & 72.0\% & 78.0\% & 54.7\% & 81.2\% & 38.6\% & 68.2\% \\ 
\textbf{LLaVA-InternLM2-20B}~\cite{2023xtuner} & 72.3\% & 78.3\% & 76.6\% & 68.2\% & 78.4\% & 46.2\% & 69.4\% \\ 
\textbf{InternLM-XComposer2*}~\cite{internlmxcomposer2} & 78.1\% & 80.4\% & 83.5\% & 73.0\% & 83.7\% & 63.6\% & 74.4\% \\ 
\shline
\rowcolor{LIGHT_RED}
\multicolumn{8}{c}{\textbf{Proprietary VLMs}} \\
\textbf{Qwen-VL-Plus}~\cite{bai2023qwen} & 64.6\% & 66.5\% & 79.1\% & 50.2\% & 73.9\% & 42.9\% & 57.8\% \\ 
\textbf{Gemini-Pro-V}~\cite{team2023gemini} & 70.2\% & 70.0\% & 78.9\% & 65.9\% & 82.9\% & 46.2\% & 65.9\% \\ 
\textbf{GPT-4v}~\cite{OpenAI2023GPT4TR} & 74.3\% & 77.6\% & 73.8\% & 71.5\% & 85.3\% & 63.6\% & 68.6\% \\ 
\textbf{Qwen-VL-Max}~\cite{bai2023qwen} & 75.4\% & 74.8\% & 87.2\% & 67.0\% & 85.3\% & 54.9\% & 70.5\% \\ 
\shline
\end{tabular}%
}
\end{table}

\noindent \textbf{Performance on MMBench-\texttt{test}. }
We first conduct a sanity check by inferring MMBench questions with GPT-4, using text-only inputs. 
After conducting the rigorous quality control paradigm in \Cref{method:collection}, 
GPT-4 demonstrates a random-level overall accuracy. 
Among open-source VLMs, InternLM-XComposer2~\cite{internlmxcomposer2} achieves the best performance and surpass other open-source or proprietary models by a large margin, \wrt the overall score, 
demonstrating its superior ability in multimodal understanding. 
After that, models adopting the architecture of LLaVA~\cite{liu2023visual} (LLaVA series and Yi-VL series) 
also showcase strong overall performance, 
which is just inferior to the state-of-the-art closed-source GPT-4v and Qwen-VL-Max.
With a small parameter size ($\le$ 3B), MiniCPM-V achieves over 60\% Top-1 accuracy, highlighting the potential of small-scale VLMs.
Models including MiniGPT, IDEFICS, VisualGLM, and InstructBLIP demonstrate significantly inferior performance compared to other VLMs, 
while OpenFlamingo v2 shows random-level performance due to the lack of instruction tuning. 

\noindent \textbf{LLM plays a vital role. }
From the evaluation results, we find that the large language model (LLM) adopted plays a vital role in the VLM performance.
For instance, all LLaVA series VLMs (v1.5-7B, v1.5-13B, InternLM2-20B) adopt the same vision backbone and are trained with the same multimodal corpus, 
while switching the LLM from Vicuna-v1.5~\cite{zheng2023judging} to the more powerful InternLM2-20B~\cite{2023internlm} leads to steady improvement across all L-2 capabilities (especially significant for reasoning tasks).
The scaling also holds for variants with different sizes from the same LLM family. 
By adopting the 13B variant of Vicuna rather than the 7B variant, 
VLMs in the MiniGPT, InstructBLIP, and LLaVA v1.5 series outperform their 7B counterparts by 8.3\%, 1.5\%, and 3.5\% overall Top-1 accuracies on the MMBench-\texttt{test} split, respectively.

\begin{figure}[t]
\vspace{-8mm}
\centering
\includegraphics[width=.95\linewidth]{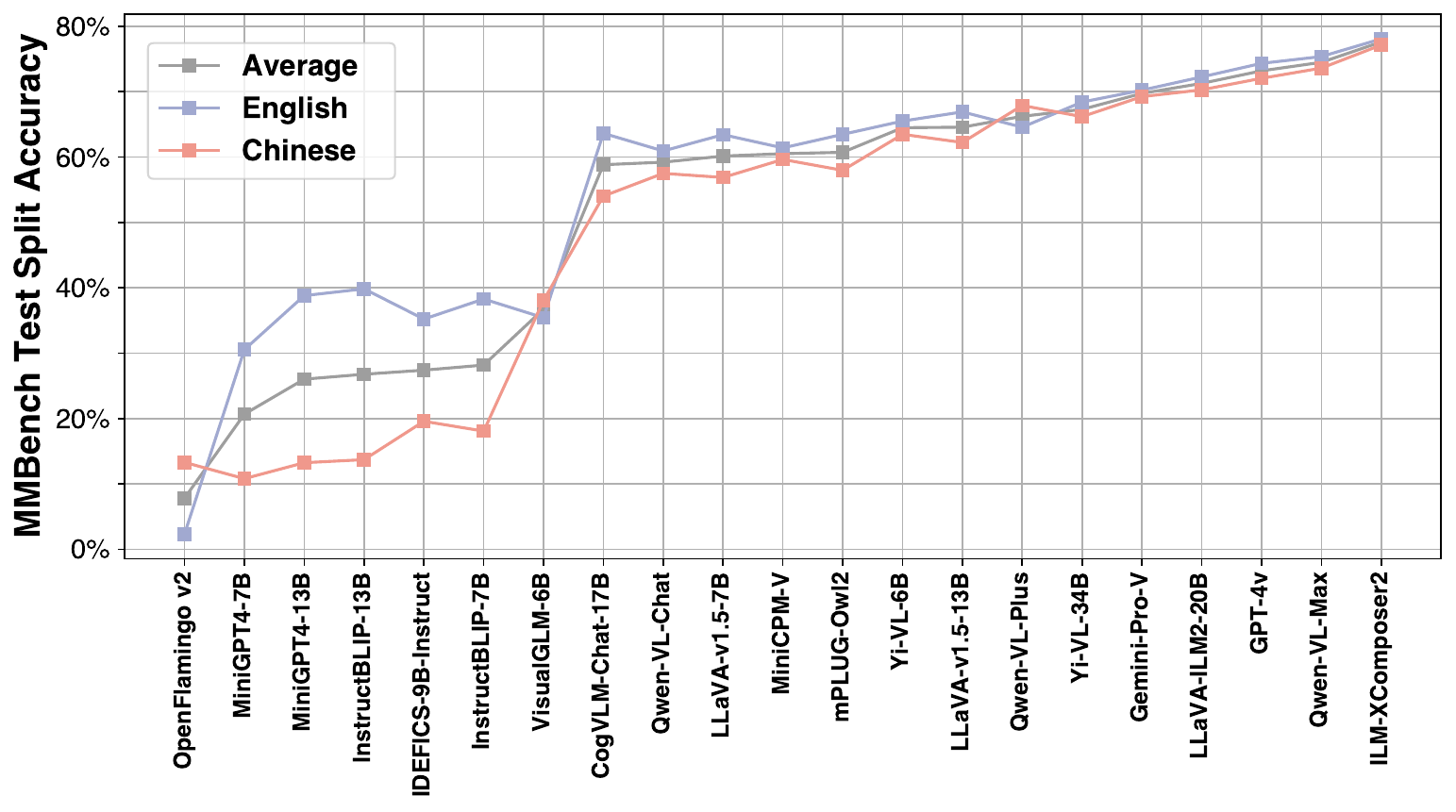}
\vspace{-2mm}
\caption{\textbf{The performance on the \texttt{test} split of MMBench and MMBench-CN. } Models are sorted with the ascending order of average performance. ILM stands for InternLM. }
\label{fig:bilingual-comparison}
\vspace{-3mm}
\end{figure}

\noindent \textbf{Performance on MMBench-CN. } 
In \Cref{fig:bilingual-comparison}, we compare the performance of different VLMs on MMBench and MMBench-CN. 
Most VLMs display a lower performance on MMBench-CN compared to the results on MMBench,
except OpenFlamingo v2, VisualGLM, and Qwen-VL-Plus.  
The difference may be attributed to the unbalanced English and Chinese corpora used in 
the pretraining and instruction-tuning of VLMs and their corresponding LLMs.
We notice that most top-performing VLMs on MMBench also display outstanding performance under the bilingual context.
The largest EN-CN performance gap for models that achieve 70+\% Top-1 accuracy on MMBench is a mere 2\%, 
For InternLM-XComposer2, the accuracy only drops by less than 1\% when evaluated on MMBench-CN. 
Such an advantage can be attributed to utilizing LLMs with better bilingual capabilities or 
tuning the VLM with more balanced cross-language multimodal corpora.

\subsection{Fine-grained Analysis}
\label{sec:analysis}

In this section, we present more fine-grained analysis based on the evaluation results. 

\noindent \textbf{Content Moderation of Proprietary VLMs. }
When we take an in-depth look at the predictions of proprietary VLMs, 
we notice that all of them apply explicit content moderation. 
GPT-4v, Gemini-Pro-V, and Qwen-VL-Max reject answering in 1.8\%, 1.6\%, and 0.1\% of cases across all CircularEval passes in MMBench, respectively.
74\% of questions rejected by GPT-4v are related to celebrity recognition (\Cref{fig:error_case}), 
while no obvious rejection pattern is observed for Gemini-Pro-V.
Under CircularEval, such moderation has a negative impact on the evaluated accuracy.
To estimate an \textbf{upper-bound} performance, we assume that VLMs can perfectly answer all rejected questions and re-calculate the accuracy. 
\Cref{tab:upperbound} shows that the content moderation policy affects the MMBench-\texttt{test} accuracy by up to 2.4\%, which is not a significant change.

\noindent
\begin{table}[t]
\begin{minipage}{0.35\linewidth}
\captionof{table}{\textbf{`Upper-bound' Acc Estimation for Proprietary VLMs. } }
\label{tab:upperbound}
\centering
\resizebox{\linewidth}{!}{
\tablestyle{8pt}{1.5}
\begin{tabular}{l|cc}
\shline
Model        & MMBench-\texttt{test}  & Upper Bound \\
\shline
GPT-4v       & 74.3 & 76.2        \\
Gemini-Pro-V & 70.2 & 72.6        \\
Qwen-VL-Max  & 75.4 & 75.5        \\ \shline
\end{tabular}}
\end{minipage}
\hfill
\begin{minipage}{0.63\linewidth}
\centering
\includegraphics[width=\linewidth]{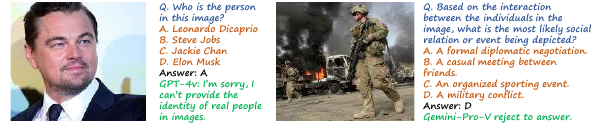}
\vspace{-3mm}
\captionof{figure}{\textbf{Content Moderation Cases of Proprietary VLMs. }}
\label{fig:error_case}
\end{minipage}
\end{table}

\begin{figure}[t]
\vspace{-3mm}
\centering
\includegraphics[width=\linewidth]{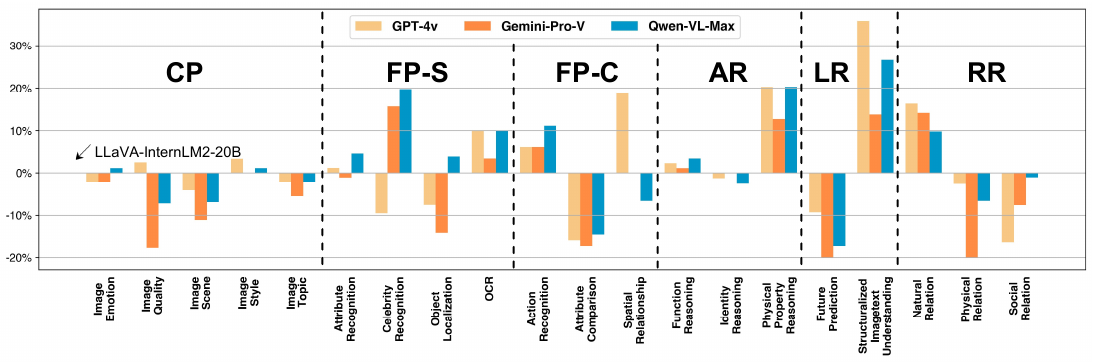}
\vspace{-2mm}
\caption{\textbf{Proprietary VLMs \emph{vs.} Open-Source ones at a fine-grained level. }  }
\label{fig:api_cmp}
\vspace{-4mm}
\end{figure}

\noindent \textbf{Proprietary \emph{vs.} Open-Source: What is the gap? }
Compared to the varied performance of open-source VLMs,
most proprietary models demonstrate competitive performance on MMBench.
This raises a question we care about: 
are proprietary models generally more powerful,
or do each kind of model display unique strengths and weaknesses across different types of ability?
To answer this question, we perform a fine-grained comparison of three proprietary VLMs and LLaVA-InternLM2-20B, 
the top-performing model trained on open-source datasets only, 
and visualize the result in \Cref{fig:api_cmp}. 
We observe that proprietary models significantly outperform the open-source ones under two major scenarios:
i) \textbf{Structuralized image-text understanding}, which requires VLMs to understand complex codes, tables, diagrams, or layouts.
ii) \textbf{Tasks requiring external knowledge to solve}, which correspond to abilities including celebrity recognition, physical property reasoning, natural relation reasoning, \emph{etc.}
Meanwhile, proprietary VLMs do not display advantages on tasks corresponding to other perception or reasoning capabilities. 

\begin{figure}[t]
\vspace{0mm}
\centering
\includegraphics[width=\linewidth]{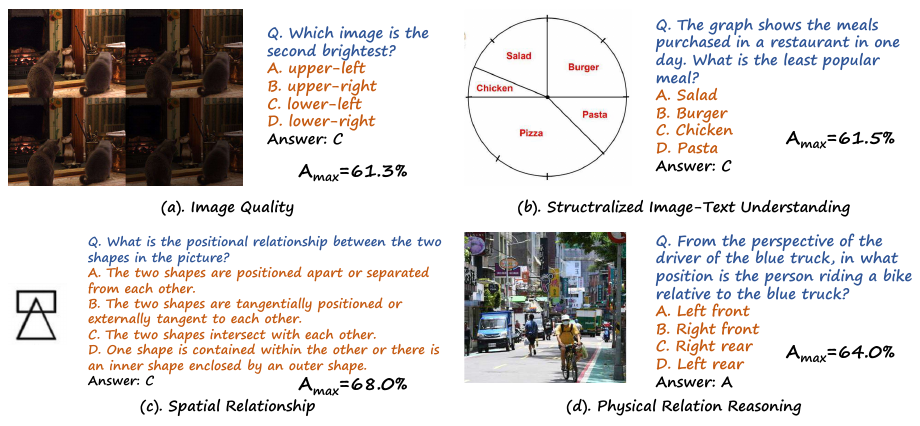}
\vspace{-3mm}
\caption{\textbf{Hard examples that belong to the 4 L-3 abilities with lowest $A_{max}$. } All VLMs have made the wrong prediction for the visualized examples under CircularEval.  }
\label{fig:hard_example}
\vspace{-3mm}
\end{figure}

\noindent \textbf{Hard cases in MMBench. }
For most VLMs, the fine-grained accuracies vary a lot across different ability categories.
To provide insights for future VLM optimization,
we find the maximum accuracy ($A_{max}$) across all evaluated VLMs on each L-3 capability.
Samples belonging to L-3 capabilities with the lowest $A_{max}$ are visualized in \Cref{fig:hard_example}. 
Generally, we find that all existing VLMs have the following limitations:
1. Poor at recognizing the low-level features on visual inputs, \emph{i.e.}, they cannot accurately recognize and compare the brightness, sharpness, contrast ratio, or artifacts of images.
2. Difficulty in understanding structuralized visual inputs like tables, diagrams, or layouts, even for relatively simple cases like \Cref{fig:hard_example}(b);
3. Perform badly on recognizing or reasoning about the inter-object spatial relationships, either in 2D or 3D space.

\section{Conclusion}
\label{sec:discuss}

We introduce MMBench, a multi-modality benchmark that performs objective evaluation for VLMs with over 3,000 multiple-choice questions covering 20 ability dimensions. 
To produce robust and reliable evaluation results, 
we introduce a new evaluation strategy named \textbf{CircularEval}.
The strategy is much stricter than the vanilla 1-pass evaluation and can yield reliable evaluation results at an affordable cost. 
Considering the limited instruction following ability of some VLMs,
to yield more accurate evaluation results, 
we additionally adopt LLMs to extract choices from the model's predictions.
We comprehensively evaluate over 20 mainstream VLMs on MMBench, covering different architectures and parameter sizes. 
The evaluation results provide valuable insights for future improvements. 


\newpage
\appendix

\section{More Details about the Data}
\label{sec:abi_def}

In this section, we begin by providing a detailed definition of each leaf ability (L-3) and present a collection of visualization samples that are directly related to each leaf ability. Then, we enumerate all the data sources that were utilized in the construction of MMBench.

\subsection{Definition about Each Leaf Ability}

\begin{figure}[ht!]
    \includegraphics[width=\textwidth]{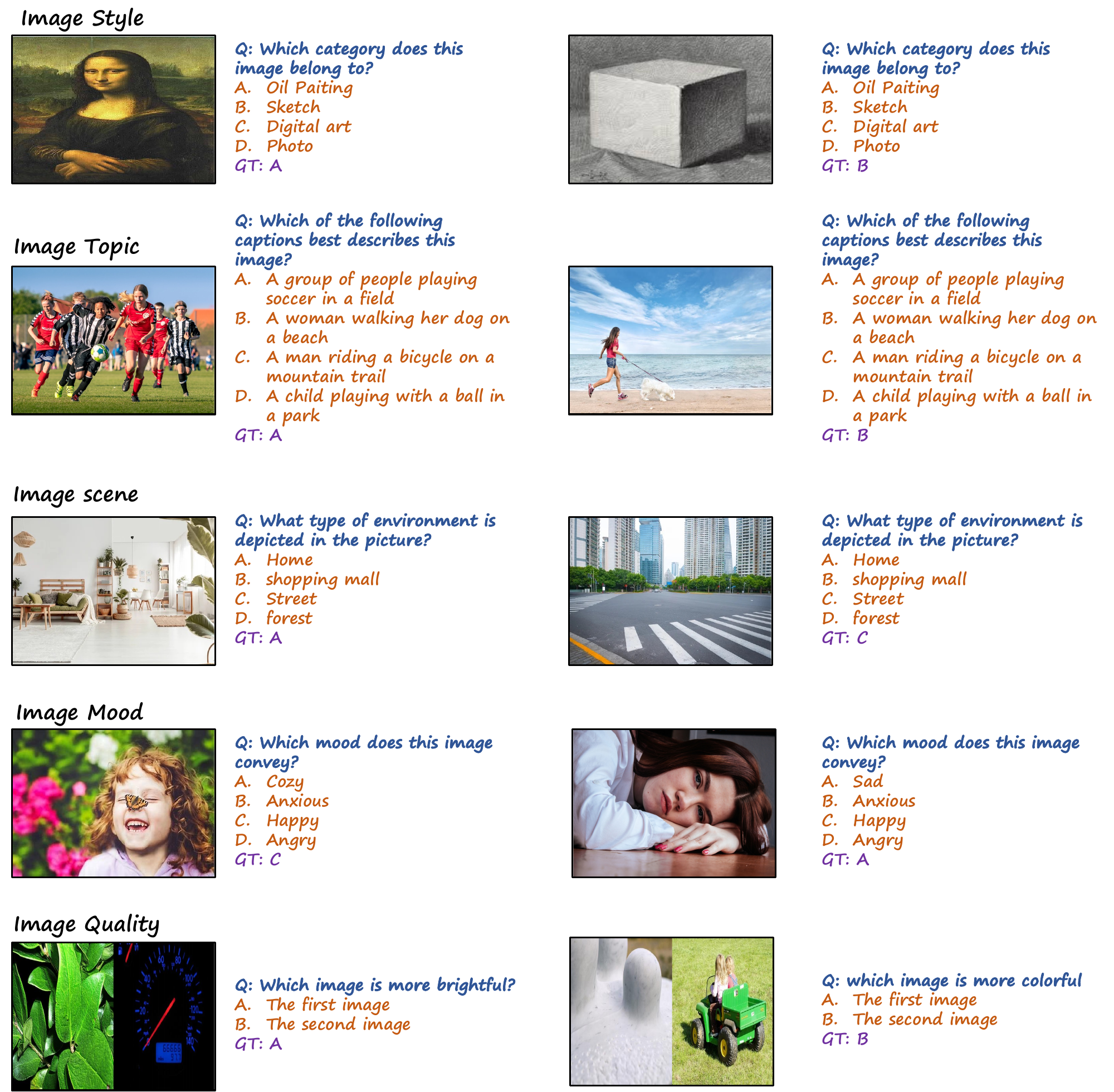}
    \caption{\textbf{Coarse Perception: Data samples. }}
    \label{fig:coarse_perception}
\end{figure}

\subsection*{Coarse Perception}
\begin{enumerate}[leftmargin=*]
    \item \textbf{Image Style}: Determine which type of image it belongs to, such as photos, paintings, CT scans, etc.
    \item \textbf{Image Scene}: Determine which environment is shown in the image, such as indoors, outdoors, forest, city, mountains, waterfront, sunny day, rainy day, etc.
    \item \textbf{Image Emotion}: Determine which subjective emotion is conveyed by the overall image, such as cold, cheerful, sad, or oppressive.
    \item \textbf{Image Quality}: Determine the objective quality of the image, such as whether it is blurry, bright or dark, contrast, etc.
    \item \textbf{Image Topic}: Determine what the subject of the image is, such as scenery, portrait, close-up of an object, text, etc.
\end{enumerate}

\noindent
In \Cref{fig:coarse_perception}, we visualize data samples belonging to the \textbf{Coarse Perception} capability. 

\begin{figure}[ht!]
    \includegraphics[width=\textwidth]{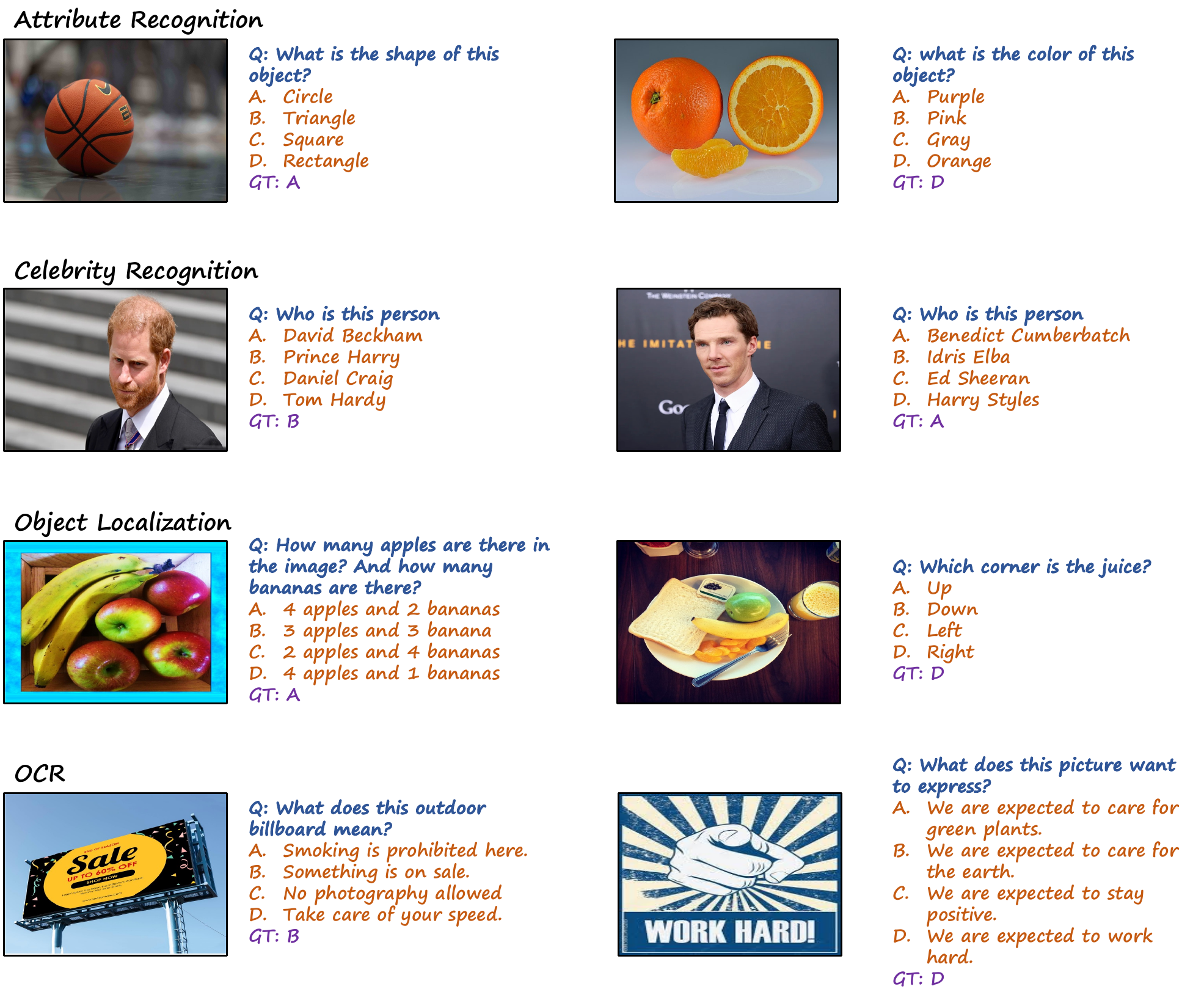}
    \caption{\textbf{Fine-grained Perception (single-instance): Data samples. }}
    \label{fig:finegrain_perception}
\end{figure}

\subsection*{Fine-grained Perception (single-instance)}
\begin{enumerate}[leftmargin=*]
    \item \textbf{Object Localization}: For a single object, determine its position in the image (such as top, bottom, etc.), its absolute coordinates in the image, count the number of objects, and the orientation of the object.
    \item \textbf{Attribute Recognition}: Recognition of texture, shape, appearance characteristics, emotions, category.
    \item \textbf{Celebrity Recognition}: Recognition of celebrities, landmarks, and well-known objects.
    \item \textbf{OCR}: Recognition of text, formula, and sheet in the image.
\end{enumerate}

\noindent
In \Cref{fig:finegrain_perception}, we visualize data samples belonging to the \textbf{Fine-grained Perception (single-instance)} capability. 

\begin{figure}[ht!]
    \includegraphics[width=\textwidth]{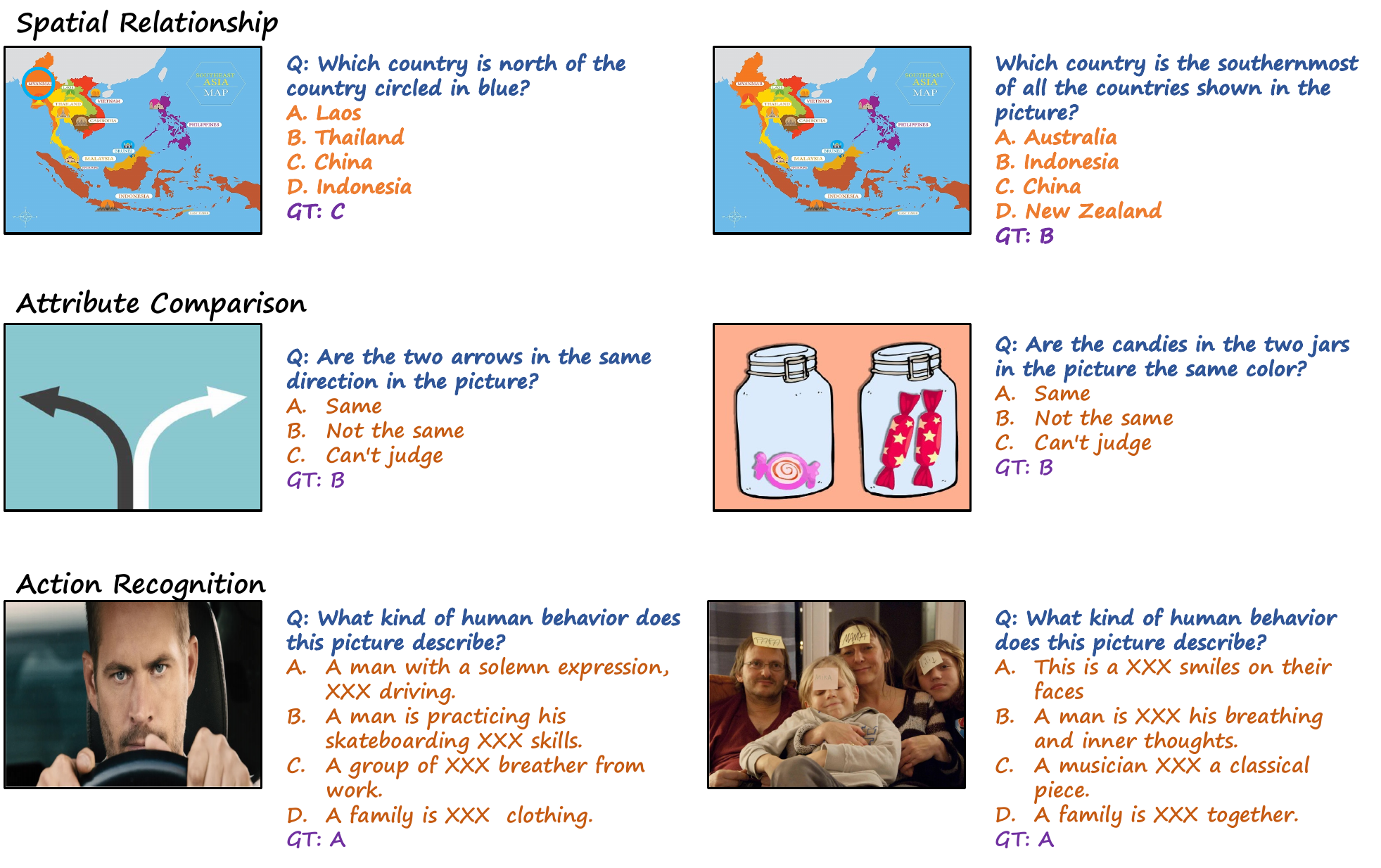}
    \caption{\textbf{Fine-grained Perception (cross-instance): Data samples. } \textbf{XXX} indicates omitted contents which are less relevant to the question. }
    \label{fig:finegrain_perception_cross}
\end{figure}

\subsection*{Fine-grained Perception (cross-instance)}
\begin{enumerate}[leftmargin=*]
    \item \textbf{Spatial Relationship}: Determine the relative position between objects in image.
    \item \textbf{Attribute Comparison}: Compare attributes of different objects in image, such as shape, color, etc.
    \item \textbf{Action Recognition}: Recognizing human actions, including pose motion, human-object interaction, and human-human interaction.
\end{enumerate}

\noindent
In \Cref{fig:finegrain_perception_cross}, we visualize data samples belonging to the \textbf{Fine-grained Perception (cross-instance)} capability.

\begin{figure}[ht!]
    \includegraphics[width=\textwidth]{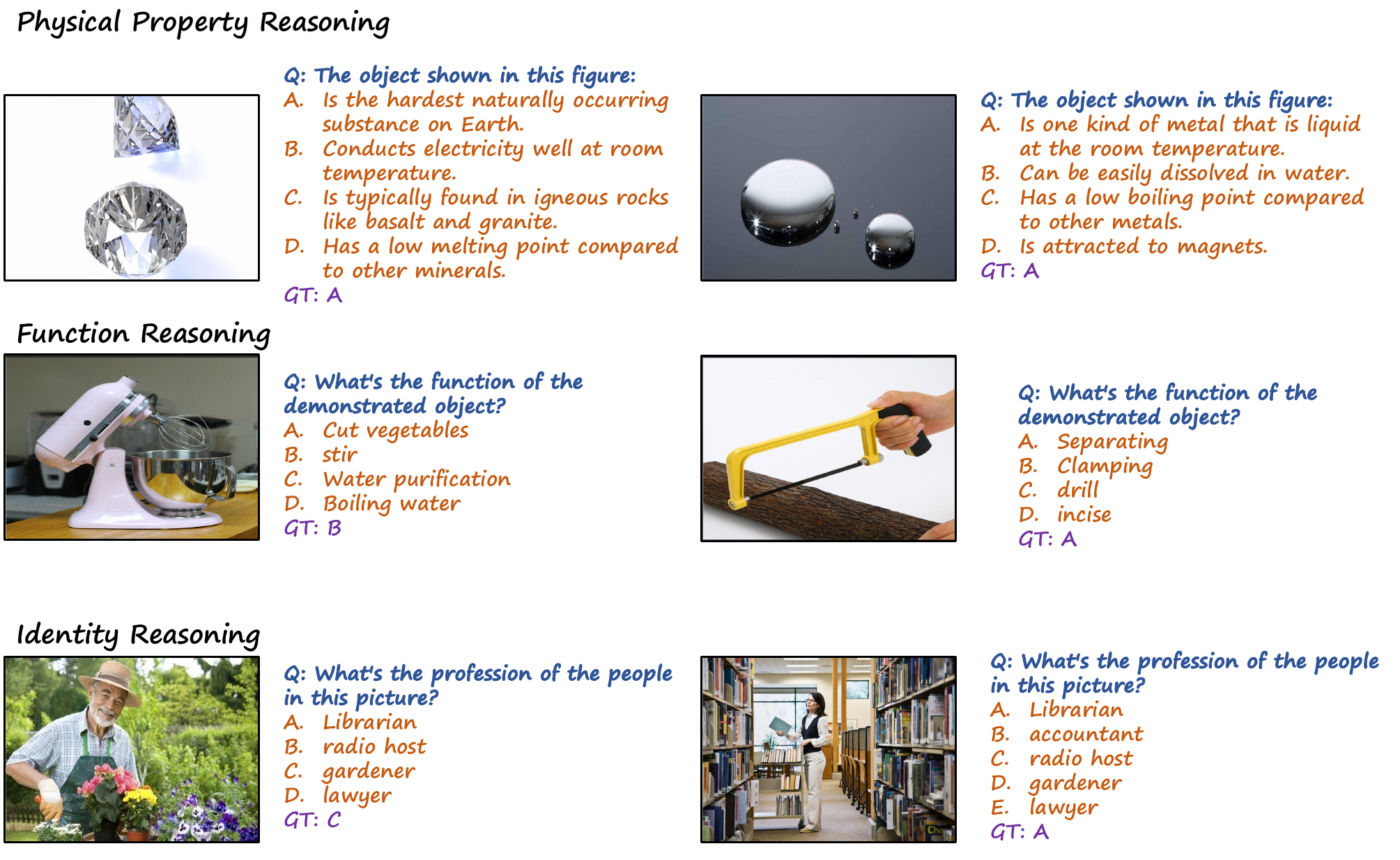}
    \caption{\textbf{Attribute Reasoning: Data samples. }}
    \label{fig:attribute_reasonning}
\end{figure}

\subsection*{Attribute Reasoning}
\begin{enumerate}[leftmargin=*]
    \item \textbf{Physical Property Reasoning}: Predict the physical property of an object. Examples: he physical property of concentrated sulfuric acid is that it is volatile, the physical property of water is its fluidity, etc.
    \item \textbf{Function Reasoning}: Predict the function of an object. Examples: the function of a broom is to sweep the floor, the function of a spatula is to cook, the function of a pen is to write, etc.
    \item \textbf{Identity Reasoning}: Predict the identity of a person. Example: by observing a person’s clothing and appearance, one may infer his / her occupation.
\end{enumerate}

\noindent
In \Cref{fig:attribute_reasonning}, we visualize data samples belonging to the \textbf{Attribute Reasoning} capability. 

\begin{figure}[ht!]
    \includegraphics[width=\textwidth]{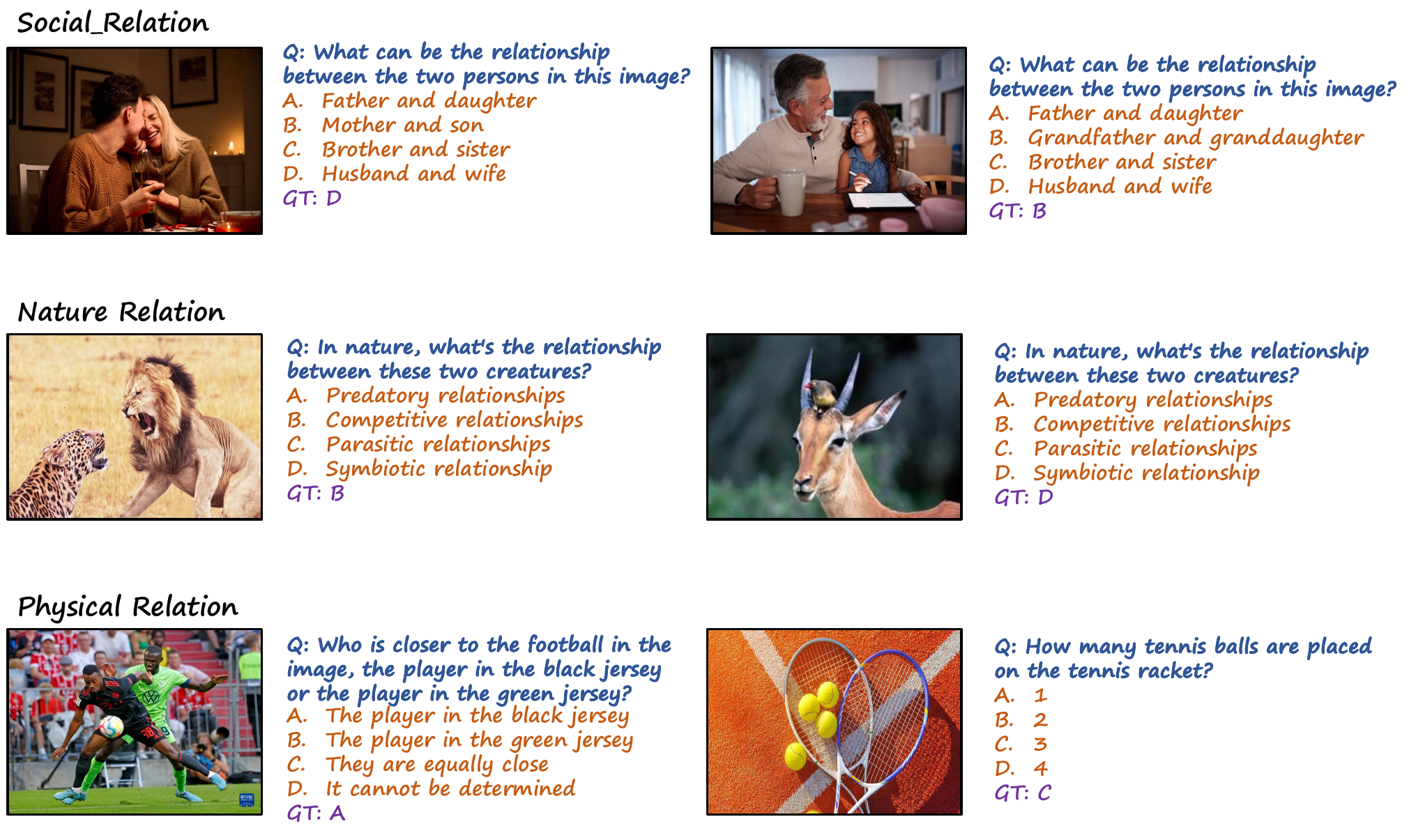}
    \caption{\textbf{Relation Reasoning: Data samples. }}
    \label{fig:relation_reasoning}
\end{figure}

\subsection*{Relation Reasoning}
\begin{enumerate}[leftmargin=*]
    \item \textbf{Social Relation}: Relations in human society or relations defined from the human perspective. Examples: Inter-person relations, such as father and son, husband and wife, friend, hostile, etc.
    \item \textbf{Physical Relation}: All relationships that exist in the physical world, 3D spatial relationships and the connections between objects are.
    \item \textbf{Nature Relation}: Other abstract relationships that exist in nature. Examples: predation, symbiosis, coexistence, etc.
\end{enumerate}

\noindent
In \Cref{fig:relation_reasoning}, we visualize data samples belonging to the \textbf{Relation Reasoning} capability.

\begin{figure}[ht!]
    \includegraphics[width=\textwidth]{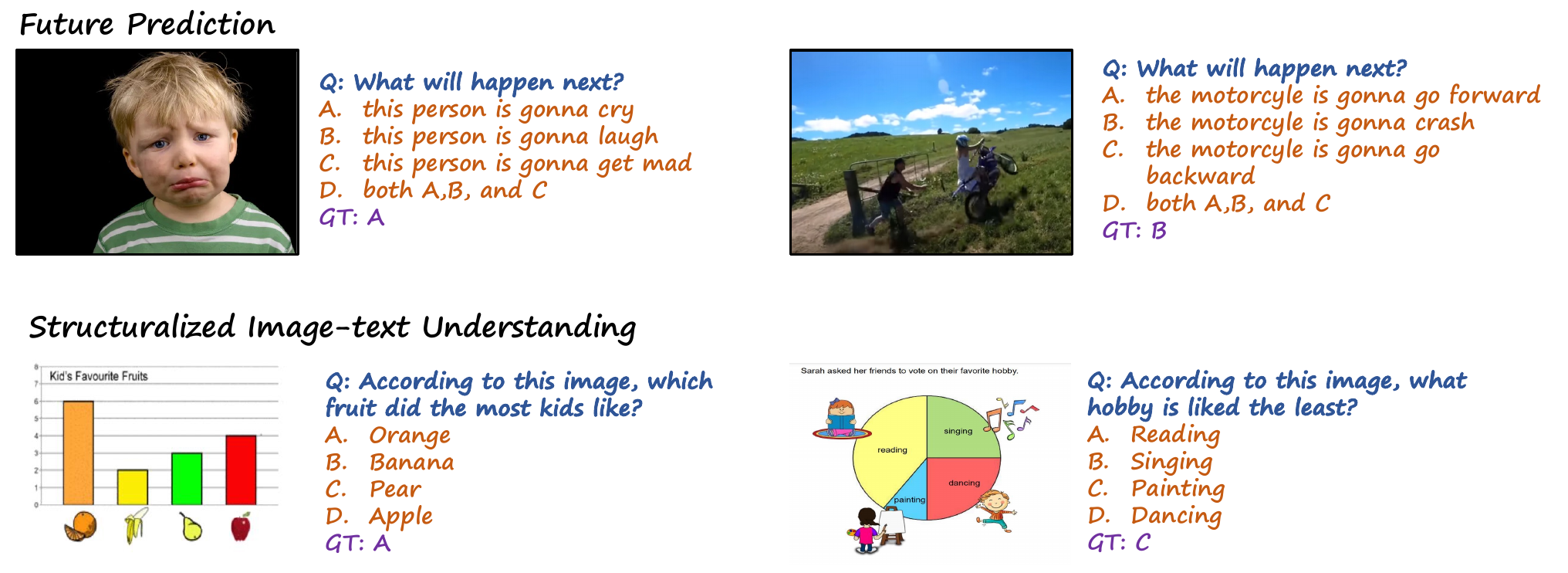}
    \caption{\textbf{Logic Reasoning: Data samples. }}
    \label{fig:logic_reasoning}
\end{figure}

\subsection*{Logic Reasoning}
\begin{enumerate}[leftmargin=*]
    \item \textbf{Structuralized Image-Text Understanding}: Structured understanding of images and text, including parsing the content of charts (such as the trends of multiple bars in a bar chart), understanding the code in an image, etc.
    \item \textbf{Future Prediction}: Predict what will happen in the future. Examples: if it is thundering in the sky now, it can be predicted that it will rain soon (physical phenomenon); if someone raises their fist, it means they are going to hit someone (event occurrence); if someone’s face becomes serious, it means they are going to get angry (emotional change).
\end{enumerate}

\noindent
In \Cref{fig:logic_reasoning}, we visualize data samples belonging to the \textbf{Logic Reasoning} capability.

\subsection{Data Sources of MMBench}

Just as we introduce in Section 3.2 of the main paper, MMBench is mainly collected from the Internnet (80\%) and the validation set of some public datasets (20\%). \Cref{tab:source} lists all these sources for images, questions and choices in MMBench.

\begin{table}[h]

\tabstyle{5pt}
\caption{
\textbf{The source of $(Q,C,I,A)$ in \ourmethod.} \textbf{Customize} means all of question, choices and answer are constructed by us. \textbf{Customize \& selection} implies that these components are either constructed by us or selected from the original dataset. } 
\vspace{2mm}
\label{tab:source}
\resizebox{.8\textwidth}{!}{
\tablestyle{15pt}{1.2}
\begin{tabular}{l c c c}
\toprule
\textbf{Image Source} & \textbf{Problem Source} & \textbf{Number}  & \textbf{Ratio} \\ 
\midrule
ARAS~\cite{ARAS} & customize \& selection &  76 & 2.4\% \\
CLEVR~\cite{johnson2017clevr} & customize \& selection & 14 & 0.4\% \\
COCO~\cite{coco_caption} & customize \& selection &  179 & 5.6\% \\
KonIQ-10k~\cite{koniq10k} & customize \& selection &  32 & 1.0\% \\
LLaVA~\cite{liu2023visual} & customize & 19 & 0.6\% \\
PISC~\cite{li2017dual} & customize \& selection & 15 & 0.5\% \\
Places~\cite{zhou2017places} & customize \& selection & 59 & 1.8\% \\
ScienceQA~\cite{scienceqa} & customize \& selection & 156 & 4.8\%  \\
ShapeWorld~\cite{kuhnle2017shapeworld} & customize \& selection & 20 & 0.6\%  \\
TextVQA~\cite{textvqa} & customize \& selection & 18 &  0.6\%  \\
VSR~\cite{Liu2022VisualSR} & customize \& selection & 19 & 0.6\% \\
W3C School~\cite{W3Cschool} & customize  &  20 & 0.6\%  \\
Internet & customize &  2590 & 80.5\% \\
\bottomrule
\end{tabular}
}
\end{table}

\section{More Details on MMBench Construction}

In this section we provide more qualitative results on the quality control paradigm we adopted to construct MMBench, as well as the prompt we used for MMBench-CN translation.

\begin{figure}[ht!]
    \includegraphics[width=\textwidth]{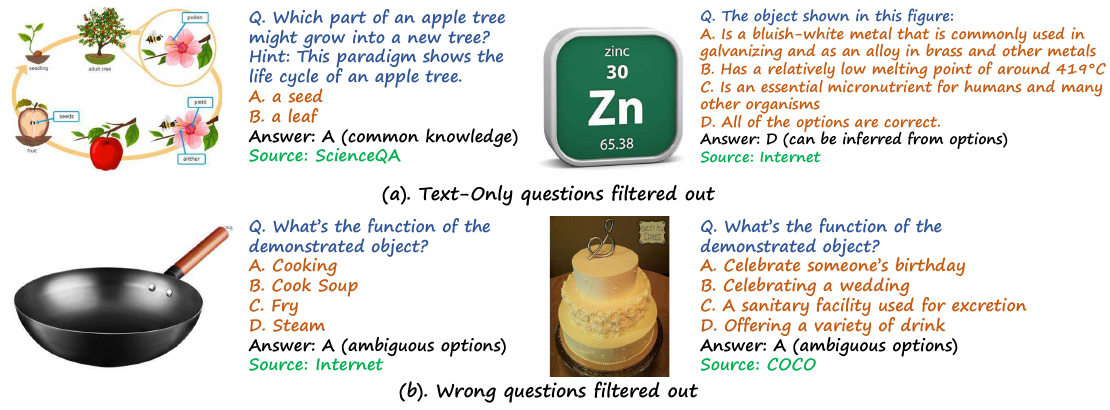}
    \caption{\textbf{Unqualified samples filtered out in MMBench. }}
    \label{fig:quality_control_sample}
\end{figure}

\noindent \textbf{\textit{`Text-only'} question filtering. }
To filter out the `\textbf{text-only}' questions (which can be answered correctly with text-only inputs by LLMs) from MMBench. 
We apply three state-of-the-art LLMs, including GPT-4~\cite{OpenAI2023GPT4TR}, Gemini-Pro~\cite{team2023gemini}, and Qwen-Max~\cite{qwen} to infer the questions with text-only inputs under CircularEval.
If more than two LLMs answer the question correctly, 
the question will be manually checked and removed if it is unqualified. 
In \Cref{fig:quality_control_sample}(a), we visualize some unqualified questions filtered out by this approach.

\noindent \textbf{\textit{`Wrong'} question filtering. }
During preliminary study, we also notice that some data samples in MMBench might be \textit{wrong}, 
due to ambiguous questions or options, 
repeated options, or incorrect answers. 
To filter out these wrong samples, we infer MMBench questions with three proprietary VLMs (GPT-4v, Gemini-Pro-V, Qwen-VL-Max) and two opensource VLMs (InternLM-XComposer2 and LLaVA-v1.5-13B). 
If no VLM can answer a question correctly under CircularEval, the question will then be manually checked. 
In \Cref{fig:quality_control_sample}(b), we visualize wrong samples filtered out by the approach.

\begin{figure}[ht!]
    \includegraphics[width=\textwidth]{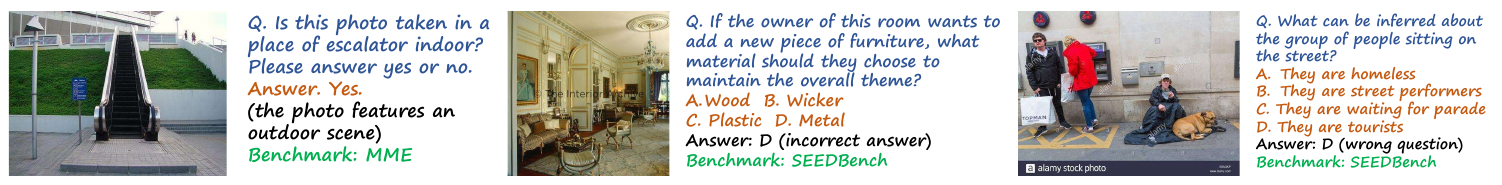}
    \caption{\textbf{Unqualified samples in other benchmarks can also be detected by our quality control paradigms. }}
    \label{fig:other_bench_wrong}
\end{figure}

\noindent \textbf{The Universality of the Quality Control Paradigm. }
The quality control paradigm adopted by MMBench is general and can also be applied to other benchmarks to improve the quality. 
To support this claim, we apply the quality control paradigm to other popular multimodal evaluation benchmarks (like MME~\cite{Fu2023MMEAC} and SEEDBench~\cite{li2023seed}) and try to detect the low-quality samples. 
We find that our quality control paradigm can also successfully detect and filter out unqualified samples from these benchmarks.
Some detected samples are visualized in \Cref{fig:other_bench_wrong}.

\noindent \textbf{MMBench-CN Translation. }
In \Cref{fig:prompt_translation}, we provide the prompt we adopted for MMBench-CN translation, which include instructions and several in-context examples. 
All translations generated by GPT-4 will be further manually verfied to ensure the correctness.

\begin{figure*}[!ht] 
\begin{AIbox}{B.1 MMBench-CN Translation}  
{
\begin{CJK*}{UTF8}{gbsn}
你是一个翻译助手，你的任务是帮我把下面的英文题目及选项翻译成中文，并保持完全一样的含义。
你仅需要翻译文本中的英文内容，不需要翻译其他语言的内容，请只翻译给定内容，不要丢失 / 修改 / 添加内容。 
对于文本中的专有名词，符号，代码，或是人名等，请依然保持英文，不需要翻译。
我会以 ``json'' 格式给出题目及选项的内容，你需要把翻译后的中文内容以 ``json'' 格式返回给我。
\\
例 1：\\
英文: \\
\{"Q": "Which of the following was part of the role of a deaconess? ", "A": "Ministering to the sick", "B": "Preparing women for baptism", "C": "Praying for the suffering"\} \\
中文: \\
\{"Q": "以下哪项是女执事的职责之一？", "A": "照顾病人", "B": "为女性准备洗礼", "C": "为受苦的人祷告"\}
\\
例 2： \\
英文: \\
\{"Q": "Which can be the associated text with this image posted on twitter? ", "A": "Located in Bome County, Nyingchi City, Tibet of China, the Yigong Iron Mountain is always surrounded by clouds and mist during summer.", "B": "夏天 所有季节中最闪耀的季节 阳光明媚，万物清明 泰山向人们展现的初夏之景 处处充满着诗情画意", "C": "Giant logs and stripped trees on Rialto Beach in the Olympic National Park.  \#beach \#wawx \#blackandwhite @yourtake", "D": "Madison Falls in Olympic National Park, WA [OC] [3024x4032] \#nature"\} \\
中文: \\
\{"Q": "与这张推特上图片配套的推文是什么？", "A": "坐落在中国西藏自治区林芝市波密县的易贡铁山，在夏季总是被云雾环绕。", "B": "夏天 所有季节中最闪耀的季节 阳光明媚，万物清明 泰山向人们展现的初夏之景 处处充满着诗情画意", "C": "奥林匹克国家 Rialto 沙滩上的巨木与被剥皮的树木。\#beach \#wawx \#blackandwhite @yourtake", "D": "Madison 瀑布，奥林匹克国家公园， WA [OC] [3024x4032] \#nature"\}
\\
请翻译： \\
英文: \\
\{The English question presented in the json format\} \\
中文: 
\end{CJK*}
}
\end{AIbox} 
\caption{\textbf{An example prompt of Chinese single choice with reasoning.}}
\label{fig:prompt_translation}
\end{figure*}

\section{More Details on LLM-based Choice Extraction}

\begin{figure}[ht!]
    \includegraphics[width=\textwidth]{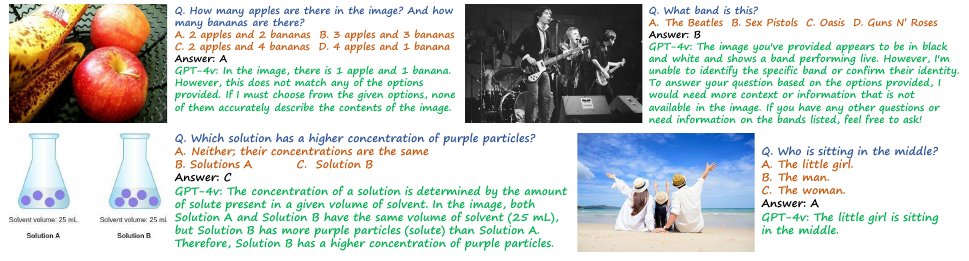}
    \caption{\textbf{Failure cases of GPT-4v during exact matching. }}
    \label{fig:exact_failure}
\end{figure}

\noindent \textbf{Failure Cases of Heuristic Matching. }
In \Cref{fig:exact_failure}, we display some failure cases of heuristic matching of the state-of-the-art VLM GPT-4v. 
Basically, such failure may occur when the VLM:
i) rejects or is not capable to answer the given question; 
ii) answers the question in different words rather than the correct choice; 
iii) provides an answer with multiple choice labels (A, B, C, \emph{etc. }) included.

\noindent \textbf{The prompt for LLM-based Choice Extraction. }
In \Cref{fig:prompt_extract}, we provide the prompt we adopted for LLM-based choice extraction. 
In-context examples are included to improve the instruction-following capability of the LLM adopted. 

\begin{figure*}[!ht] 
\begin{AIbox}{C.1 Prompt for Choice Extraction}  
{
\begin{CJK*}{UTF8}{gbsn}
You are an AI assistant who will help me to match an answer with several options of a single-choice question. 
You are provided with a question, several options, and an answer, and you need to find which option is most similar to the answer. 
If the meaning of all options are significantly different from the answer, output Z. 
You should only do the matching based exactly on the literal meaning of the options and answer. 
You should not perform any external inference based on your knowledge during the matching. 
Your should output a single uppercase character in A, B, C, D (if they are valid options), and Z.  \\
Example 1: \\
Question: What is the main object in image? \\
Options: A. teddy bear B. rabbit C. cat D. dog \\
Answer: a cute teddy bear \\
Your output: A \\
Example 2: \\
Question: What is the main object in image? \\
Options: A. teddy bear B. rabbit C. cat D. dog \\
Answer: Spider \\
Your output: Z \\
Now it's your turn: \\
Question: \{question\} \\
Options: \{options\} \\ 
Answer: \{answer\} \\
Your output: 
\end{CJK*}
}
\end{AIbox} 
\caption{\textbf{The prompt used for choice extraction on MMBench.} The Chinese translation of this prompt is adopted for MMBench-CN choice extraction. }
\label{fig:prompt_extract}
\end{figure*}

\noindent \textbf{Performance Evaluated with Other Choice Extractors. }
In \Cref{tab:extractor_abl}, we list the MMBench-\texttt{dev} performance obtained with different choice extractors, 
including GPT-4 (0125), GPT-3.5-Turbo (0613 and 0125), and InternLM2-7B~\cite{2023internlm}.
VLMs with high success rate (>99\%) in heuristic matching are skipped. 
From the table, we see that adopting different choice extractors will not lead to significant different evaluation results. 
VisualGLM displays the largest range across all choice extractors, which is around 1.4\%. 
For top-performing proprietary VLMs (GPT-4v, Gemini-Pro-V, \emph{etc.}), the gap is at most 0.3\%.

\begin{table}[t]
\centering
\caption{\textbf{MMBench-\texttt{dev} accuracies with different choice extractors under \textit{CircularEval}. }} 
\label{tab:extractor_abl}
\resizebox{\textwidth}{!}{%
\tablestyle{6pt}{1.5}
\begin{tabular}{lccccc}
\shline
\rowcolor{COLOR_MEAN}
\textbf{VLM} & \textbf{\multilinecell{Exact\\Matching}} & \textbf{\multilinecell{GPT-4-Turbo\\(0125)}} & \textbf{\multilinecell{GPT-3.5-Turbo\\(0613)}} & \textbf{\multilinecell{GPT-3.5-Turbo\\(0125)}} & \textbf{InternLM2-7B} \\ 
\shline
\textbf{MiniGPT4-7B}~\cite{zhu2023minigpt} & 26.0 & 32.7 & 33.1 & 33.0 & 32.9 \\ 
\textbf{IDEFICS-9B-Instruct}~\cite{laurencon2023obelics} & 36.0 & 37.2 & 37.2 & 37.2 & 37.2 \\ 
\textbf{InstructBLIP-7B}~\cite{dai2023instructblip} & 34.8 & 37.4 & 37.5 & 37.5 & 37.7 \\ 
\textbf{VisualGLM-6B}~\cite{du2022glm} & 19.4 & 36.1 & 37.5 & 37.5 & 36.1 \\ 
\textbf{MiniGPT4-13B}~\cite{zhu2023minigpt} & 30.7 & 37.5 & 37.8 & 37.8 & 37.6 \\ 
\textbf{InstructBLIP-13B}~\cite{dai2023instructblip} & 36.6 & 40.9 & 41.1 & 41.0 & 41.4 \\ 
\textbf{Qwen-VL-Chat}~\cite{bai2023qwen} & 56.7 & 59.5 & 59.8 & 59.4 & 59.8 \\ 
\textbf{Qwen-VL-Plus}~\cite{bai2023qwen} & 43.7 & 62.9 & 62.6 & 61.9 & 63.2 \\ 
\textbf{MiniCPM-V}~\cite{minicpm2024} & 57.6 & 64.8 & 64.7 & 64.6 & 64.7 \\ 
\textbf{Gemini-Pro-V}~\cite{team2023gemini} & 70.4 & 70.9 & 70.9 & 70.9 & 70.8 \\ 
\textbf{GPT-4v}~\cite{OpenAI2023GPT4TR} & 71.8 & 74.3 & 74.6 & 74.6 & 74.6 \\ 
\textbf{Qwen-VL-Max}~\cite{bai2023qwen} & 72.9 & 76.4 & 76.5 & 76.2 & 76.5 \\  
\shline
\end{tabular}%
}
\end{table}

\begin{table}[ht]
\tabstyle{10pt}
\caption{\textbf{LLM-based Matching $vs$ Exact Matching. } A preliminary study on VQA benchmarks. 
\textbf{Accuracy} is the success rate of answers being exactly matched with the groundtruth. 
For each sample, \textbf{GPT score} is an integer $n \in [1, 5]$, indicating the similarity between answer and groundtruth.
We adopt \textbf{gpt-3.5-turbo-0613} as the matching LLM and report the average GPT score for testing samples.}
\label{tab:gpt-vqa}
\resizebox{\textwidth}{!}{
\tablestyle{6pt}{1.3}
\begin{tabular}{l|cc|cc|cc}
\shline
\rowcolor{COLOR_MEAN}
Dataset   & \multicolumn{2}{c|}{GQA~\cite{hudson2019gqa}} & \multicolumn{2}{c|}{OK-VQA~\cite{ok-vqa}} & \multicolumn{2}{c}{Text-VQA~\cite{textvqa}} \\ 
\shline
Model     & Flamingo   & MiniGPT-4  & Flamingo    & MiniGPT-4    & Flamingo     & MiniGPT-4     \\ 
\shline
Accuracy  & 33.6\%     & 22.4\%     & 42.6\%      & 21.9\%       & 22.9\%       & 9.8\%         \\
Average GPT score & 2.75       & 2.74       & 2.79        & 1.97         & 1.92         & 1.54          \\ \shline
\end{tabular}}
\vspace{-5pt}
\end{table}

\begin{table}[ht]
\caption{MiniGPT-4 GQA answers (different from GT) that can be matched by ChatGPT-based evaluation. }
\label{tab:gqa-examples}
\resizebox{\textwidth}{!}{
\tablestyle{4pt}{1.3}
\begin{tabular}{l|cc|c@{}}
\toprule
\rowcolor{COLOR_MEAN}
\textbf{Question} & \textbf{GroundTruth} & \textbf{MiniGPT-4 Answer} & \textbf{GPT Score} \\ \midrule
On which side of the picture is the leather bag? & right & leather bag is on right side & 5 \\ 
Who is wearing a shirt?	& girl & woman is wearing shirt & 5 \\
What is the pen made of? & plastic & pen is made of plastic\#\# & 5 \\
What does the male person ride on? & bike & bicycle & 4 \\
Is the microwave silver and rectangular? & Yes & microwave is silver and rectangular & 5 \\ 
How does the silver lamp appear to be, on or off? & off & silver lamp appears to be off \#\# & 5 \\
\bottomrule
\end{tabular}}
\end{table}

\noindent
\textbf{LLM-based sementic matching is generally helpful. }
To demonstrate that LLMs can be a general tool for semantic matching, 
we also validate the LLM-involved evaluation paradigm on existing multi-modality tasks, 
including GQA~\cite{hudson2019gqa}, OK-VQA~\cite{ok-vqa}, and Text-VQA~\cite{textvqa}. 
Given the ground-truth answer, we use GPT-3.5-Turbo to measure the similarity between VLM's prediction\footnote{The simlarity score is an integer in [1, 5]. 
1 means completely wrong, while 5 means completely correct. }.
For each benchmark, we randomly select 1000 testing samples and evaluate with exact match (the traditional paradigm) and ChatGPT-based match, respectively, and list the results in \Cref{tab:gpt-vqa}. 
Basically, ChatGPT-based evaluation demonstrates the same trend compared to the exact-match accuracy on all tasks.
On GQA, two algorithms demonstrate very close performance under ChatGPT-based evaluation. 
In further investigation, we find the reason is that ChatGPT succeeds in matching slightly different answers (compared to GT) generated by MiniGPT-4, while exact matching fails (examples in \Cref{tab:gqa-examples}).
\section{Evaluation Settings and Results}

In Section 5.2 of the main paper, we give the results of different models on the \textbf{test} split of MMBench and MMBench-CN. 
In this section, we will introduce the detailed evaluation setting on MMBench,
and provide more evaluation results. 

\subsection{Evaluation Settings}

Unless stated otherwise, all results presented in this paper adhere to the conventional \textbf{zero-shot} evaluation setting. 
We have also attempted to assess these models with few-shot and chain-of-thought evaluations. 
\textbf{However, no encouraging results are observed.}
Below we provide the prompt we used for evaluating a VLM under the zero-shot setting on MMBench.

\begin{figure*}[!ht] 
\begin{AIbox}{D.1 Prompt Template for Zero-shot Inference. } 
{
\begin{CJK*}{UTF8}{gbsn}
Hint: xxx [optional] \\
Question: xxx \\ 
A. xxx \\ 
B. xxx \\
C. xxx [optional] \\
D. xxx [optional] \\
Please select the correct answer from the options above.
\end{CJK*}
}
\end{AIbox} 
\caption{\textbf{The prompt template adopted for zero-shot inference. }}
\label{fig:zero_shot_prompt}
\end{figure*}

\subsection{Model Settings}

In Table~\ref{tab:model_details}, we provide details of all open-source models evaluated in MMBench, including several additional models that do not fit the space of the main article. 

\begin{table}[t]
\centering
\caption{\textbf{Details of the evaluated Open-Source VLMs.}}
\label{tab:model_details}
\resizebox{\textwidth}{!}{%
\tablestyle{6pt}{1.5}
\begin{tabular}{l|llc}
\toprule
\rowcolor{COLOR_MEAN}
\textbf{VLM} & \textbf{Language Backbone} & \textbf{Vision Backbone} & \textbf{Overall Parameters}  \\ \midrule
\textbf{OpenFlamingov2}\cite{alayrac2022flamingo} & MPT 7B & CLIP ViT-L/14 & 9B \\  
\textbf{MiniGPT-4-7B}\cite{zhu2023minigpt} & Vicuna 7B & EVA-G & 8B \\ 
\textbf{IDEFICS-9B-Instruct}\cite{laurencon2023obelics} & LLaMA 7B & CLIP ViT-H/14 & 9B \\ 
\textbf{VisualGLM-6B}\cite{du2022glm} & ChatGLM 6B & EVA-CLIP & 7B  \\  
\textbf{InstructBLIP-7B}\cite{dai2023instructblip} & Vicuna 7B & EVA-G & 8B  \\ 
\textbf{MiniGPT-4-13B}\cite{zhu2023minigpt} & Vicuna 13B & EVA-G & 14B   \\ 
\textbf{PandaGPT}\cite{su2023pandagpt} & Vicuna 13B & ImageBind ViT-H/14 & 14B \\  
\textbf{InstructBLIP-13B}\cite{dai2023instructblip} & Vicuna 13B & EVA-G & 14B  \\ 
\textbf{IDEFICS-80B-Instruct}~\cite{laurencon2023obelics} & LLaMA 65B & CLIP ViT-H/14 & 80B \\ 
\textbf{Qwen-VL-Chat}\cite{bai2023qwen} & Qwen 7B & ViT-G/16 &  10B \\ 
\textbf{MiniCPM-V}\cite{minicpm2024} & MiniCPM 2.4B & SigLip-400M & 3B  \\ 
\textbf{LLaVA-v1.5-7B}\cite{liu2023improved} & Vicuna 7B & CLIP ViT-L/14 & 7B  \\ 
\textbf{mPLUG-Owl2}\cite{ye2023mplug} & LLaMA2 7B & CLIP ViT-L/14 & 8B \\ 
\textbf{CogVLM-Chat-17B}\cite{Wang2023CogVLMVE} & Vicuna 7B & EVA2-CLIP-E & 18B  \\ 
\textbf{ShareGPT4V-7B}\cite{chen2023sharegpt4v} & Vicuna 7B & CLIP ViT-L/14 & 7B  \\ 
\textbf{Yi-VL-6B}\cite{2023YiVL} & Yi-6B & CLIP ViT-H/14 & 7B  \\ 
\textbf{LLaVA-InternLM-7B}\cite{2023xtuner} & InternLM 7B & CLIP ViT-L/14 &  9B \\ 
\textbf{ShareGPT4V-13B}\cite{chen2023sharegpt4v} & Vicuna 13B & CLIP ViT-L/14 & 13B  \\ 
\textbf{LLaVA-v1.5-13B}\cite{liu2023improved} & Vicuna 13B & CLIP ViT-L/14 & 13B  \\ 
\textbf{Yi-VL-34B}\cite{2023YiVL} & Yi 34B & CLIP ViT-H/14 & 35B \\ 
\textbf{OmniLMM-12B}\cite{omnilmm2024} & Zephyr-7B-$\beta$ & EVA-02-5B & 12B \\ 
\textbf{Monkey-Chat}\cite{li2023monkey} & Qwen 7B & ViT BigG & 10B \\ 
\textbf{InternLM-XComposer}\cite{zhang2023internlm} & InternLM-7B & EVA-G &  9B \\ 
\textbf{LLaVA-InternLM2-7B}\cite{2023xtuner} & InternLM2-7B & CLIP ViT-L/14 & 9B  \\ 
\textbf{LLaVA-InternLM2-20B}\cite{2023xtuner} & InternLM2-20B & CLIP ViT-L/14 & 23B  \\ 
\textbf{InternLM-XComposer2}\cite{internlmxcomposer2} & InternLM2-7B & CLIP ViT-L/14 & 9B  \\ 
\bottomrule
\end{tabular}%
}
\end{table}

\subsection{More Results}

In this section, we give more detailed results about the performance of different models on MMBench and MMBench-CN. 
We present the detailed evaluation results of 30 different VLMs (some of them do not appear in the main paper due to limited space). 
For detailed results on each L-3 ability, see the separate sheet in the supplementary materials.

\begin{table}[t]
\centering
\caption{\textbf{CircularEval results on MMBench-\texttt{dev} set (L-2 abilities).} 
Open-source models tagged with * incorporate in-house data in model training. } 

\label{tab:main_en_dev}
\resizebox{\textwidth}{!}{%
\tablestyle{8pt}{1.5}
\begin{tabular}{lccccccc}
\shline
\textbf{Model} & \textbf{Overall} & \textbf{CP} & \textbf{FP-S} & \textbf{FP-C} & \textbf{AR} & \textbf{LR} & \textbf{RR} \\ \shline 
\rowcolor{LIGHT_BLUE}
\multicolumn{8}{c}{\textbf{OpenSource VLMs}} \\
\textbf{OpenFlamingo v2}~\cite{alayrac2022flamingo} & 2.6\% & 0.8\% & 4.5\% & 1.1\% & 5.5\% & 0.0\% & 3.4\% \\ 
\textbf{MiniGPT4-7B}~\cite{zhu2023minigpt} & 32.7\% & 38.4\% & 39.1\% & 20.7\% & 49.4\% & 10.5\% & 22.4\% \\ 
\textbf{VisualGLM-6B}~\cite{du2022glm} & 36.1\% & 40.3\% & 43.3\% & 19.6\% & 49.4\% & 16.9\% & 33.9\% \\ 
\textbf{IDEFICS-9B-Instruct}~\cite{laurencon2023obelics} & 37.2\% & 50.6\% & 37.7\% & 30.2\% & 51.8\% & 4.8\% & 25.3\% \\ 
\textbf{InstructBLIP-7B}~\cite{dai2023instructblip} & 37.4\% & 46.4\% & 47.1\% & 23.5\% & 51.2\% & 8.1\% & 24.7\% \\ 
\textbf{MiniGPT4-13B}~\cite{zhu2023minigpt} & 37.5\% & 44.2\% & 48.4\% & 16.8\% & 57.3\% & 6.5\% & 30.5\% \\ 
\textbf{InstructBLIP-13B}~\cite{dai2023instructblip} & 40.9\% & 48.6\% & 52.2\% & 18.4\% & 56.7\% & 5.6\% & 39.7\% \\ 
\textbf{PandaGPT}~\cite{su2023pandagpt} & 41.6\% & 56.1\% & 34.6\% & 34.6\% & 53.7\% & 13.7\% & 38.5\% \\ 
\textbf{IDEFICS-80B-Instruct}~\cite{laurencon2023obelics} & 42.3\% & 54.7\% & 48.1\% & 24.6\% & 57.3\% & 8.9\% & 34.5\% \\ 
\textbf{Qwen-VL-Chat*}~\cite{bai2023qwen} & 59.5\% & 70.7\% & 69.9\% & 49.7\% & 69.5\% & 25.0\% & 44.3\% \\ 
\textbf{CogVLM-Chat-17B}~\cite{Wang2023CogVLMVE} & 62.4\% & 69.6\% & 70.6\% & 56.4\% & 67.1\% & 29.0\% & 59.2\% \\ 
\textbf{LLaVA-v1.5-7B}~\cite{liu2023improved} & 62.5\% & 71.3\% & 70.6\% & 55.9\% & 70.7\% & 25.8\% & 55.7\% \\ 
\textbf{mPLUG-Owl2}~\cite{Ye2023mPLUGOwl2RM} & 63.5\% & 72.9\% & 70.2\% & 53.6\% & 70.7\% & 29.8\% & 60.3\% \\ 
\textbf{MiniCPM-V}~\cite{minicpm2024} & 64.8\% & 71.0\% & 75.1\% & 52.5\% & 72.0\% & 30.6\% & 64.9\% \\ 
\textbf{Yi-VL-6B*}~\cite{2023YiVL} & 65.6\% & 72.7\% & 73.7\% & 54.7\% & 73.2\% & 32.3\% & 65.5\% \\ 
\textbf{ShareGPT4V-7B}~\cite{chen2023sharegpt4v} & 66.2\% & 77.3\% & 75.1\% & 57.5\% & 68.3\% & 25.8\% & 63.8\% \\ 
\textbf{ShareGPT4V-13B}~\cite{chen2023sharegpt4v} & 67.0\% & 75.1\% & 77.9\% & 58.1\% & 68.9\% & 35.5\% & 61.5\% \\ 
\textbf{LLaVA-InternLM-7B}~\cite{2023xtuner} & 67.0\% & 75.7\% & 72.7\% & 57.5\% & 71.3\% & 37.1\% & 66.7\% \\ 
\textbf{LLaVA-v1.5-13B}~\cite{liu2023improved} & 67.2\% & 74.0\% & 75.1\% & 59.2\% & 68.9\% & 38.7\% & 66.7\% \\ 
\textbf{Yi-VL-34B*}~\cite{2023YiVL} & 68.2\% & 75.7\% & 73.0\% & 55.9\% & 75.6\% & 39.5\% & 70.7\% \\ 
\textbf{Monkey-Chat}~\cite{li2023monkey} & 68.8\% & 72.9\% & 79.2\% & 58.1\% & 79.3\% & 42.7\% & 62.6\% \\ 
\textbf{OmniLMM-12B*}~\cite{omnilmm2024} & 69.7\% & 75.1\% & 79.6\% & 61.5\% & 73.8\% & 37.1\% & 69.5\% \\ 
\textbf{LLaVA-InternLM2-7B}~\cite{2023xtuner} & 71.6\% & 79.8\% & 77.2\% & 62.0\% & 74.4\% & 41.1\% & 74.1\% \\ 
\textbf{LLaVA-InternLM2-20B}~\cite{2023xtuner} & 72.8\% & 80.1\% & 75.1\% & 68.2\% & 73.8\% & 46.0\% & 76.4\% \\ 
\textbf{InternLM-XComposer*}~\cite{zhang2023internlm} & 73.9\% & 79.6\% & 81.7\% & 65.4\% & 84.8\% & 39.5\% & 72.4\% \\ 
\textbf{InternLM-XComposer2*}~\cite{internlmxcomposer2} & 79.1\% & 83.4\% & 84.4\% & 68.7\% & 83.5\% & 58.1\% & 82.8\% \\ 
\shline
\rowcolor{LIGHT_RED}
\multicolumn{8}{c}{\textbf{Proprietary VLMs}} \\
\textbf{Qwen-VL-Plus}~\cite{bai2023qwen} & 62.9\% & 67.1\% & 78.9\% & 53.1\% & 71.3\% & 28.2\% & 54.6\% \\ 
\textbf{Gemini-Pro-V}~\cite{team2023gemini} & 70.9\% & 71.3\% & 81.7\% & 62.0\% & 78.7\% & 47.6\% & 70.7\% \\ 
\textbf{GPT-4v}~\cite{OpenAI2023GPT4TR} & 74.3\% & 78.5\% & 72.3\% & 66.5\% & 82.9\% & 67.7\% & 73.6\% \\ 
\textbf{Qwen-VL-Max}~\cite{bai2023qwen} & 76.4\% & 76.2\% & 87.2\% & 69.3\% & 78.7\% & 55.6\% & 78.7\% \\ 
\shline
\end{tabular}%
}
\end{table}

\begin{table}[t]
\centering
\caption{\textbf{CircularEval results on MMBench-\texttt{test} set (L-2 abilities).} 
Open-source models tagged with * incorporate in-house data in model training. } 

\label{tab:main_en_test}
\resizebox{\textwidth}{!}{%
\tablestyle{8pt}{1.5}
\begin{tabular}{lccccccc}
\shline
\textbf{Model} & \textbf{Overall} & \textbf{CP} & \textbf{FP-S} & \textbf{FP-C} & \textbf{AR} & \textbf{LR} & \textbf{RR} \\ \shline 
\rowcolor{LIGHT_BLUE}
\multicolumn{8}{c}{\textbf{OpenSource VLMs}} \\
\textbf{OpenFlamingo v2}~\cite{alayrac2022flamingo} & 2.3\% & 1.1\% & 3.5\% & 1.5\% & 5.3\% & 0.0\% & 2.7\% \\ 
\textbf{MiniGPT4-7B}~\cite{zhu2023minigpt} & 30.5\% & 37.0\% & 31.8\% & 17.2\% & 49.8\% & 9.2\% & 25.6\% \\ 
\textbf{IDEFICS-9B-Instruct}~\cite{laurencon2023obelics} & 35.2\% & 48.3\% & 31.3\% & 29.6\% & 47.8\% & 11.4\% & 25.2\% \\ 
\textbf{VisualGLM-6B}~\cite{du2022glm} & 35.4\% & 40.2\% & 38.5\% & 26.2\% & 47.8\% & 19.6\% & 29.5\% \\ 
\textbf{InstructBLIP-7B}~\cite{dai2023instructblip} & 38.3\% & 46.7\% & 39.0\% & 31.8\% & 55.5\% & 8.7\% & 31.0\% \\ 
\textbf{MiniGPT4-13B}~\cite{zhu2023minigpt} & 38.8\% & 44.6\% & 42.9\% & 23.2\% & 64.9\% & 8.2\% & 32.9\% \\ 
\textbf{PandaGPT}~\cite{su2023pandagpt} & 39.7\% & 51.9\% & 29.5\% & 27.3\% & 62.0\% & 19.0\% & 38.0\% \\ 
\textbf{InstructBLIP-13B}~\cite{dai2023instructblip} & 39.8\% & 47.2\% & 42.9\% & 21.0\% & 60.4\% & 12.5\% & 38.8\% \\ 
\textbf{IDEFICS-80B-Instruct}~\cite{laurencon2023obelics} & 40.9\% & 54.6\% & 38.1\% & 29.6\% & 52.7\% & 16.8\% & 34.9\% \\ 
\textbf{Qwen-VL-Chat*}~\cite{bai2023qwen} & 60.9\% & 68.5\% & 67.7\% & 50.2\% & 78.0\% & 37.0\% & 45.7\% \\ 
\textbf{MiniCPM-V}~\cite{minicpm2024} & 61.4\% & 65.6\% & 69.4\% & 51.3\% & 70.6\% & 35.3\% & 59.7\% \\ 
\textbf{LLaVA-v1.5-7B}~\cite{liu2023improved} & 63.4\% & 70.0\% & 68.0\% & 57.7\% & 77.6\% & 33.2\% & 56.2\% \\ 
\textbf{mPLUG-Owl2}~\cite{Ye2023mPLUGOwl2RM} & 63.5\% & 68.1\% & 69.1\% & 55.8\% & 78.4\% & 37.0\% & 57.0\% \\ 
\textbf{CogVLM-Chat-17B}~\cite{Wang2023CogVLMVE} & 63.6\% & 72.8\% & 66.6\% & 55.4\% & 71.4\% & 33.7\% & 62.0\% \\ 
\textbf{ShareGPT4V-7B}~\cite{chen2023sharegpt4v} & 64.6\% & 72.2\% & 68.7\% & 59.6\% & 72.7\% & 34.8\% & 60.5\% \\ 
\textbf{Yi-VL-6B*}~\cite{2023YiVL} & 65.5\% & 72.8\% & 72.9\% & 56.2\% & 75.5\% & 41.3\% & 55.4\% \\ 
\textbf{LLaVA-InternLM-7B}~\cite{2023xtuner} & 65.9\% & 72.6\% & 68.7\% & 57.3\% & 80.0\% & 37.5\% & 63.2\% \\ 
\textbf{ShareGPT4V-13B}~\cite{chen2023sharegpt4v} & 66.7\% & 75.6\% & 73.5\% & 56.9\% & 72.7\% & 37.0\% & 62.4\% \\ 
\textbf{LLaVA-v1.5-13B}~\cite{liu2023improved} & 66.9\% & 73.1\% & 72.4\% & 60.3\% & 75.5\% & 35.9\% & 65.5\% \\ 
\textbf{Yi-VL-34B*}~\cite{2023YiVL} & 68.4\% & 72.0\% & 78.0\% & 54.7\% & 81.2\% & 38.6\% & 68.2\% \\ 
\textbf{OmniLMM-12B*}~\cite{omnilmm2024} & 69.2\% & 72.0\% & 79.8\% & 61.0\% & 78.0\% & 40.2\% & 66.7\% \\ 
\textbf{Monkey-Chat}~\cite{li2023monkey} & 69.6\% & 75.0\% & 75.4\% & 63.3\% & 82.4\% & 46.7\% & 58.9\% \\ 
\textbf{InternLM-XComposer*}~\cite{zhang2023internlm} & 71.3\% & 75.7\% & 76.3\% & 60.3\% & 84.5\% & 44.6\% & 71.7\% \\ 
\textbf{LLaVA-InternLM2-7B}~\cite{2023xtuner} & 71.6\% & 78.1\% & 75.4\% & 66.7\% & 77.6\% & 44.6\% & 70.2\% \\ 
\textbf{LLaVA-InternLM2-20B}~\cite{2023xtuner} & 72.3\% & 78.3\% & 76.6\% & 68.2\% & 78.4\% & 46.2\% & 69.4\% \\ 
\textbf{InternLM-XComposer2*}~\cite{internlmxcomposer2} & 78.1\% & 80.4\% & 83.5\% & 73.0\% & 83.7\% & 63.6\% & 74.4\% \\ 
\shline
\rowcolor{LIGHT_RED}
\multicolumn{8}{c}{\textbf{Proprietary VLMs}} \\
\textbf{Qwen-VL-Plus}~\cite{bai2023qwen} & 64.6\% & 66.5\% & 79.1\% & 50.2\% & 73.9\% & 42.9\% & 57.8\% \\ 
\textbf{Gemini-Pro-V}~\cite{team2023gemini} & 70.2\% & 70.0\% & 78.9\% & 65.9\% & 82.9\% & 46.2\% & 65.9\% \\ 
\textbf{GPT-4v}~\cite{OpenAI2023GPT4TR} & 74.3\% & 77.6\% & 73.8\% & 71.5\% & 85.3\% & 63.6\% & 68.6\% \\ 
\textbf{Qwen-VL-Max}~\cite{bai2023qwen} & 75.4\% & 74.8\% & 87.2\% & 67.0\% & 85.3\% & 54.9\% & 70.5\% \\ 
\shline
\end{tabular}%
}
\end{table}

\begin{table}[t]
\centering
\caption{\textbf{CircularEval results on MMBench-CN-\texttt{dev} set (L-2 abilities).} 
Open-source models tagged with * incorporate in-house data in model training. } 

\label{tab:main_cn_dev}
\resizebox{\textwidth}{!}{%
\tablestyle{8pt}{1.5}
\begin{tabular}{lccccccc}
\shline
\textbf{Model} & \textbf{Overall} & \textbf{CP} & \textbf{FP-S} & \textbf{FP-C} & \textbf{AR} & \textbf{LR} & \textbf{RR} \\ \shline 
\rowcolor{LIGHT_BLUE}
\multicolumn{8}{c}{\textbf{OpenSource VLMs}} \\
\textbf{MiniGPT4-13B}~\cite{zhu2023minigpt} & 11.8\% & 14.6\% & 13.8\% & 14.0\% & 15.9\% & 3.2\% & 2.3\% \\ 
\textbf{MiniGPT4-7B}~\cite{zhu2023minigpt} & 11.9\% & 11.9\% & 14.5\% & 7.8\% & 19.5\% & 3.2\% & 10.9\% \\ 
\textbf{OpenFlamingo v2}~\cite{alayrac2022flamingo} & 14.3\% & 14.4\% & 14.9\% & 11.2\% & 21.3\% & 10.5\% & 12.6\% \\ 
\textbf{InstructBLIP-13B}~\cite{dai2023instructblip} & 15.1\% & 16.0\% & 14.9\% & 7.8\% & 30.5\% & 4.0\% & 14.4\% \\ 
\textbf{InstructBLIP-7B}~\cite{dai2023instructblip} & 18.1\% & 16.0\% & 16.6\% & 10.6\% & 38.4\% & 4.0\% & 23.6\% \\ 
\textbf{IDEFICS-9B-Instruct}~\cite{laurencon2023obelics} & 18.7\% & 22.7\% & 19.7\% & 7.3\% & 35.4\% & 1.6\% & 17.2\% \\ 
\textbf{IDEFICS-80B-Instruct}~\cite{laurencon2023obelics} & 29.2\% & 32.0\% & 27.0\% & 25.1\% & 50.0\% & 8.1\% & 26.4\% \\ 
\textbf{PandaGPT}~\cite{su2023pandagpt} & 31.0\% & 40.1\% & 24.9\% & 18.4\% & 47.6\% & 12.1\% & 33.3\% \\ 
\textbf{VisualGLM-6B}~\cite{du2022glm} & 40.6\% & 45.3\% & 48.1\% & 30.7\% & 54.3\% & 8.9\% & 37.9\% \\ 
\textbf{CogVLM-Chat-17B}~\cite{Wang2023CogVLMVE} & 52.9\% & 63.5\% & 56.4\% & 41.9\% & 65.9\% & 16.9\% & 50.0\% \\ 
\textbf{LLaVA-v1.5-7B}~\cite{liu2023improved} & 57.0\% & 69.3\% & 59.9\% & 47.5\% & 62.8\% & 25.0\% & 54.0\% \\ 
\textbf{Qwen-VL-Chat*}~\cite{bai2023qwen} & 57.6\% & 66.6\% & 68.5\% & 43.6\% & 70.1\% & 21.8\% & 48.9\% \\ 
\textbf{mPLUG-Owl2}~\cite{Ye2023mPLUGOwl2RM} & 58.1\% & 68.8\% & 65.1\% & 43.0\% & 68.9\% & 29.8\% & 50.0\% \\ 
\textbf{ShareGPT4V-7B}~\cite{chen2023sharegpt4v} & 59.7\% & 71.8\% & 62.6\% & 48.6\% & 62.8\% & 26.6\% & 61.5\% \\ 
\textbf{OmniLMM-12B*}~\cite{omnilmm2024} & 60.6\% & 67.7\% & 69.9\% & 48.0\% & 70.1\% & 25.8\% & 59.2\% \\ 
\textbf{ShareGPT4V-13B}~\cite{chen2023sharegpt4v} & 62.4\% & 72.9\% & 67.1\% & 55.3\% & 66.5\% & 34.7\% & 55.7\% \\ 
\textbf{LLaVA-v1.5-13B}~\cite{liu2023improved} & 62.5\% & 71.8\% & 65.7\% & 57.0\% & 67.1\% & 33.1\% & 59.8\% \\ 
\textbf{MiniCPM-V}~\cite{minicpm2024} & 63.0\% & 68.2\% & 75.1\% & 53.1\% & 72.0\% & 25.8\% & 60.3\% \\ 
\textbf{LLaVA-InternLM-7B}~\cite{2023xtuner} & 63.0\% & 72.4\% & 68.2\% & 50.3\% & 68.9\% & 35.5\% & 62.1\% \\ 
\textbf{Monkey-Chat}~\cite{li2023monkey} & 65.1\% & 73.8\% & 74.4\% & 50.3\% & 77.4\% & 37.9\% & 54.6\% \\ 
\textbf{Yi-VL-6B*}~\cite{2023YiVL} & 65.3\% & 72.4\% & 73.0\% & 53.1\% & 70.7\% & 33.9\% & 67.8\% \\ 
\textbf{Yi-VL-34B*}~\cite{2023YiVL} & 67.0\% & 73.8\% & 73.0\% & 52.5\% & 72.6\% & 40.3\% & 71.8\% \\ 
\textbf{LLaVA-InternLM2-7B}~\cite{2023xtuner} & 70.0\% & 81.5\% & 72.3\% & 59.2\% & 73.8\% & 34.7\% & 74.7\% \\ 
\textbf{InternLM-XComposer*}~\cite{zhang2023internlm} & 71.3\% & 76.5\% & 77.5\% & 63.7\% & 81.7\% & 37.9\% & 71.8\% \\ 
\textbf{LLaVA-InternLM2-20B}~\cite{2023xtuner} & 71.7\% & 77.9\% & 74.4\% & 68.7\% & 75.6\% & 43.5\% & 74.1\% \\ 
\textbf{InternLM-XComposer2*}~\cite{internlmxcomposer2} & 77.2\% & 83.4\% & 84.1\% & 64.2\% & 84.1\% & 54.8\% & 75.9\% \\ 
\shline
\rowcolor{LIGHT_RED}
\multicolumn{8}{c}{\textbf{Proprietary VLMs}} \\
\textbf{Qwen-VL-Plus}~\cite{bai2023qwen} & 67.5\% & 68.8\% & 83.0\% & 54.2\% & 75.6\% & 38.7\% & 65.5\% \\ 
\textbf{Gemini-Pro-V}~\cite{team2023gemini} & 69.3\% & 72.4\% & 78.5\% & 63.1\% & 78.7\% & 40.3\% & 65.5\% \\ 
\textbf{GPT-4v}~\cite{OpenAI2023GPT4TR} & 73.3\% & 76.5\% & 71.6\% & 67.0\% & 82.3\% & 63.7\% & 74.1\% \\ 
\textbf{Qwen-VL-Max}~\cite{bai2023qwen} & 75.9\% & 73.8\% & 85.8\% & 71.5\% & 81.7\% & 55.6\% & 77.0\% \\ 
\shline
\end{tabular}%
}
\end{table}

\begin{table}[t]
\centering
\caption{\textbf{CircularEval results on MMBench-CN-\texttt{test} set (L-2 abilities).} 
Open-source models tagged with * incorporate in-house data in model training. } 

\label{tab:main_cn_test}
\resizebox{\textwidth}{!}{%
\tablestyle{8pt}{1.5}
\begin{tabular}{lccccccc}
\shline
\textbf{Model} & \textbf{Overall} & \textbf{CP} & \textbf{FP-S} & \textbf{FP-C} & \textbf{AR} & \textbf{LR} & \textbf{RR} \\ \shline 
\rowcolor{LIGHT_BLUE}
\multicolumn{8}{c}{\textbf{OpenSource VLMs}} \\
\textbf{MiniGPT4-7B}~\cite{zhu2023minigpt} & 10.8\% & 9.4\% & 11.8\% & 5.6\% & 24.5\% & 4.9\% & 8.5\% \\ 
\textbf{MiniGPT4-13B}~\cite{zhu2023minigpt} & 13.2\% & 16.3\% & 13.5\% & 9.0\% & 27.3\% & 3.8\% & 4.3\% \\ 
\textbf{OpenFlamingo v2}~\cite{alayrac2022flamingo} & 13.3\% & 16.5\% & 10.2\% & 9.0\% & 18.8\% & 11.4\% & 12.4\% \\ 
\textbf{InstructBLIP-13B}~\cite{dai2023instructblip} & 13.7\% & 13.7\% & 14.6\% & 6.4\% & 26.5\% & 4.3\% & 14.3\% \\ 
\textbf{InstructBLIP-7B}~\cite{dai2023instructblip} & 18.1\% & 15.7\% & 18.6\% & 9.4\% & 31.4\% & 8.7\% & 25.2\% \\ 
\textbf{IDEFICS-9B-Instruct}~\cite{laurencon2023obelics} & 19.6\% & 22.4\% & 17.4\% & 7.1\% & 35.9\% & 6.0\% & 24.4\% \\ 
\textbf{IDEFICS-80B-Instruct}~\cite{laurencon2023obelics} & 28.8\% & 33.0\% & 26.9\% & 25.1\% & 41.2\% & 13.6\% & 26.0\% \\ 
\textbf{PandaGPT}~\cite{su2023pandagpt} & 29.6\% & 40.4\% & 20.0\% & 12.0\% & 49.8\% & 13.0\% & 34.1\% \\ 
\textbf{VisualGLM-6B}~\cite{du2022glm} & 38.1\% & 44.8\% & 39.4\% & 22.8\% & 55.5\% & 18.5\% & 34.9\% \\ 
\textbf{CogVLM-Chat-17B}~\cite{Wang2023CogVLMVE} & 54.0\% & 66.1\% & 49.7\% & 47.6\% & 67.8\% & 26.1\% & 49.6\% \\ 
\textbf{LLaVA-v1.5-7B}~\cite{liu2023improved} & 56.9\% & 65.2\% & 53.6\% & 52.1\% & 75.5\% & 31.0\% & 50.8\% \\ 
\textbf{Qwen-VL-Chat*}~\cite{bai2023qwen} & 57.5\% & 63.0\% & 64.5\% & 41.6\% & 74.7\% & 35.9\% & 50.0\% \\ 
\textbf{mPLUG-Owl2}~\cite{Ye2023mPLUGOwl2RM} & 58.0\% & 64.4\% & 57.1\% & 50.2\% & 75.1\% & 31.5\% & 56.6\% \\ 
\textbf{ShareGPT4V-7B}~\cite{chen2023sharegpt4v} & 58.3\% & 67.2\% & 58.2\% & 51.3\% & 72.7\% & 28.3\% & 54.7\% \\ 
\textbf{MiniCPM-V}~\cite{minicpm2024} & 59.6\% & 64.8\% & 66.6\% & 52.8\% & 69.0\% & 33.2\% & 54.3\% \\ 
\textbf{OmniLMM-12B*}~\cite{omnilmm2024} & 60.8\% & 64.8\% & 66.4\% & 53.9\% & 74.7\% & 30.4\% & 58.9\% \\ 
\textbf{LLaVA-v1.5-13B}~\cite{liu2023improved} & 62.2\% & 68.3\% & 61.5\% & 56.9\% & 73.5\% & 35.9\% & 64.3\% \\ 
\textbf{ShareGPT4V-13B}~\cite{chen2023sharegpt4v} & 62.7\% & 69.6\% & 63.6\% & 56.2\% & 74.7\% & 36.4\% & 60.9\% \\ 
\textbf{Yi-VL-6B*}~\cite{2023YiVL} & 63.5\% & 68.7\% & 71.7\% & 52.4\% & 74.7\% & 39.7\% & 56.6\% \\ 
\textbf{LLaVA-InternLM-7B}~\cite{2023xtuner} & 64.1\% & 70.7\% & 63.8\% & 55.8\% & 75.5\% & 39.7\% & 65.5\% \\ 
\textbf{Monkey-Chat}~\cite{li2023monkey} & 65.0\% & 71.5\% & 68.9\% & 52.1\% & 80.0\% & 46.7\% & 57.4\% \\ 
\textbf{Yi-VL-34B*}~\cite{2023YiVL} & 66.2\% & 69.6\% & 75.6\% & 56.2\% & 80.0\% & 37.0\% & 61.2\% \\ 
\textbf{InternLM-XComposer*}~\cite{zhang2023internlm} & 69.2\% & 74.8\% & 71.7\% & 58.1\% & 80.8\% & 39.1\% & 75.6\% \\ 
\textbf{LLaVA-InternLM2-7B}~\cite{2023xtuner} & 69.9\% & 75.4\% & 72.9\% & 63.7\% & 81.2\% & 42.4\% & 68.6\% \\ 
\textbf{LLaVA-InternLM2-20B}~\cite{2023xtuner} & 70.3\% & 75.6\% & 73.5\% & 67.4\% & 75.1\% & 46.2\% & 69.4\% \\ 
\textbf{InternLM-XComposer2*}~\cite{internlmxcomposer2} & 77.1\% & 80.4\% & 82.8\% & 71.2\% & 88.2\% & 55.4\% & 72.1\% \\ 
\shline
\rowcolor{LIGHT_RED}
\multicolumn{8}{c}{\textbf{Proprietary VLMs}} \\
\textbf{Qwen-VL-Plus}~\cite{bai2023qwen} & 67.9\% & 69.6\% & 78.4\% & 60.3\% & 75.1\% & 48.9\% & 61.2\% \\ 
\textbf{Gemini-Pro-V}~\cite{team2023gemini} & 69.2\% & 68.1\% & 77.3\% & 64.0\% & 80.4\% & 45.7\% & 69.8\% \\ 
\textbf{GPT-4v}~\cite{OpenAI2023GPT4TR} & 72.1\% & 75.0\% & 70.1\% & 70.0\% & 82.4\% & 60.9\% & 69.4\% \\ 
\textbf{Qwen-VL-Max}~\cite{bai2023qwen} & 73.6\% & 74.4\% & 82.6\% & 69.3\% & 79.2\% & 55.4\% & 69.0\% \\ 
\shline
\end{tabular}%
}
\end{table}

\clearpage
\bibliography{main}
\bibliographystyle{plain}
\end{document}